\documentclass[10pt,twocolumn,letterpaper]{article}

\usepackage{amsmath,amsfonts,bm}

\def\eqref#1{equation~\ref{#1}}

\def\1{\bm{1}}

\DeclareMathAlphabet{\mathsfit}{\encodingdefault}{\sfdefault}{m}{sl}
\SetMathAlphabet{\mathsfit}{bold}{\encodingdefault}{\sfdefault}{bx}{n}

\newcommand{\R}{\mathbb{R}}

\usepackage{cvpr}
\usepackage{times}
\usepackage{epsfig}
\usepackage{graphicx}
\usepackage{amsmath}
\usepackage{amssymb}
\usepackage{multicol}
\usepackage{multirow}
\usepackage{booktabs}
\usepackage{bbding}
\usepackage{makecell}
\usepackage{subcaption}
\usepackage{caption}
\usepackage{color}
\definecolor{gray}{RGB}{200, 200, 200}

\usepackage[breaklinks=true,bookmarks=false]{hyperref}

\begin{document}
\title{Not All Frame Features Are Equal: Video-to-4D Generation via Decoupling Dynamic-Static Features}

\author{Liying Yang$^{1}$, Chen Liu$^{2}$, Zhenwei Zhu$^{1}$, Ajian Liu$^{3}$, Hui Ma$^{1}$, Jian Nong$^{1}$, Yanyan Liang$^{1\dagger}$\\
$^{1}$ Macau University of Science and Technology \qquad $^{2}$ The University of Queensland \\
\qquad $^{3}$ Institute of Automation, Chinese Academy of Sciences (CASIA) \\
$^{\dagger}$ Corresponding Author\\
{\tt\small lyyang69@gmail.com \qquad yyliang@must.edu.mo}}

\maketitle
\begin{abstract} 
Recently, the generation of dynamic 3D objects from a video has shown impressive results. Existing methods directly optimize Gaussians using whole information in frames. However, when dynamic regions are interwoven with static regions within frames, particularly if the static regions account for a large proportion, existing methods often overlook information in dynamic regions and are prone to overfitting on static regions. This leads to producing results with blurry textures. We consider that decoupling dynamic-static features to enhance dynamic representations can alleviate this issue. Thus, we propose a dynamic-static feature decoupling module (DSFD). Along temporal axes, it regards the regions of current frame features that possess significant differences relative to reference frame features as dynamic features. Conversely, the remaining parts are the static features. Then, we acquire decoupled features driven by dynamic features and current frame features. Moreover, to further enhance the dynamic representation of decoupled features from different viewpoints and ensure accurate motion prediction, we design a temporal-spatial similarity fusion module (TSSF). Along spatial axes, it adaptively selects similar information of dynamic regions. Hinging on the above, we construct a novel approach, DS4D. Experimental results verify our method achieves state-of-the-art (SOTA) results in video-to-4D. In addition, the experiments on a real-world scenario dataset demonstrate its effectiveness on the 4D scene. Our code will be publicly available.

\end{abstract}

\section{Introduction} \label{intro}

\begin{figure}[t]
\centering
\includegraphics[width=0.4\textwidth]{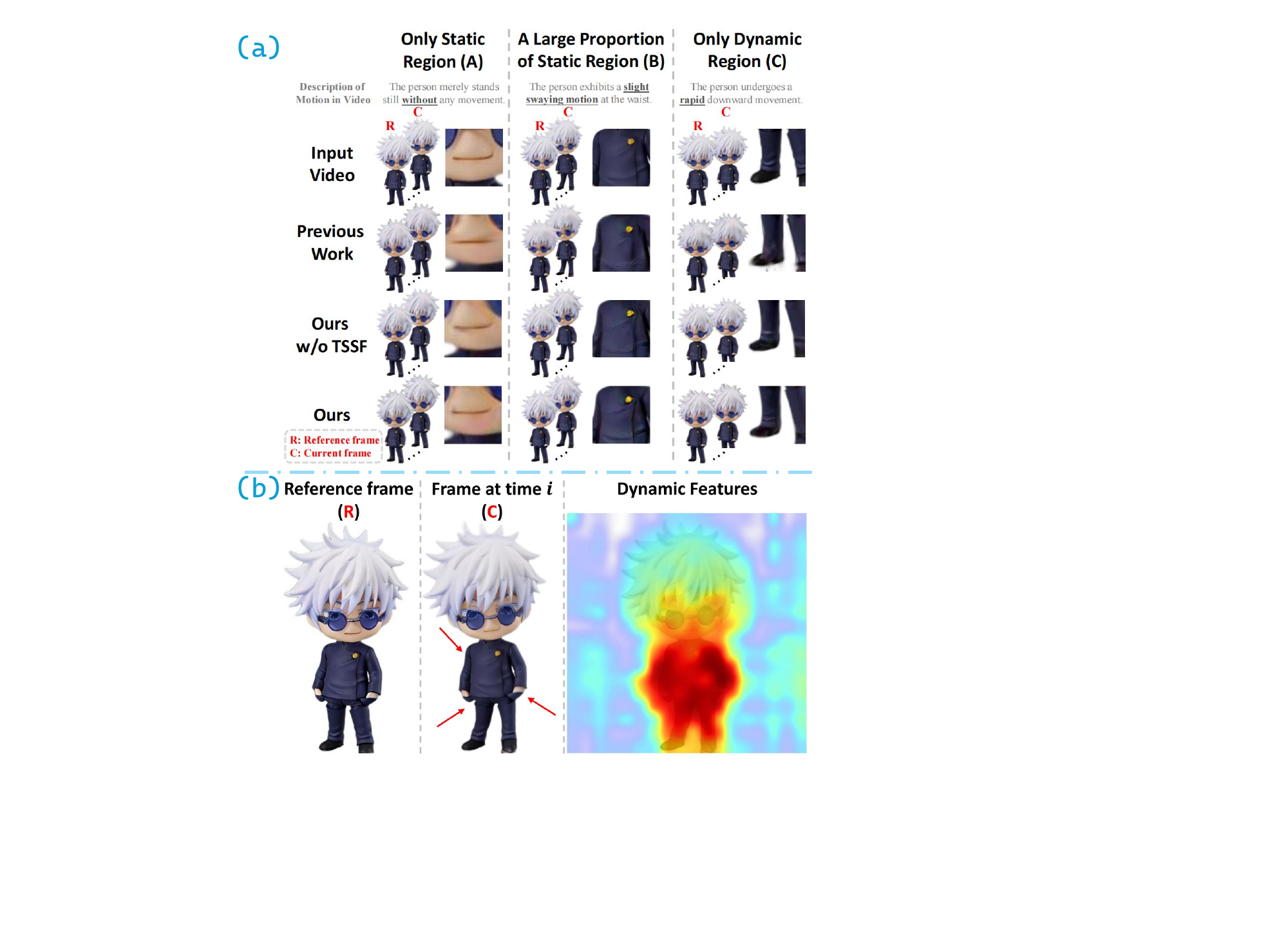}
\caption{\textbf{(a) Illustration of the issues caused by different proportions of dynamic and static regions.} Previous work \cite{zeng2024stag4d} generates the 4D content with obviously blurry textures in the dynamic regions with B-type video input. In contrast, our methods with decoupling dynamic-static features generates high-quality results with clear textures. \textbf{(b) The visualization of dynamic features in our method.} The red region highlights the primary region of interest in the dynamic features in B-type videos. This part is also the dynamic region between frames R and C. It demonstrates our method can successfully decouple dynamic-static features.}\label{highlight}
\end{figure}

Generating dynamic 3D (4D) content  \cite{li2022neural,shao2023tensor4d,li2021neural,pumarola2021d,park2021hypernerf,gao2022monocular,li2024dreammesh4d,wu20244d} from video is an essential research topic involving the field of computer vision and computer graphics. However, it is rather formidable to predict accurate motion from a few viewpoints while ensuring high-quality generation.

There are two main streams in current approaches to improving generation quality. Inference-based methods \cite{xie2024sv4d,zhang20244diffusion,ren2024l4gm,liang2024diffusion4d} can generate high-quality 4D content by capturing temporal-spatial correlations in its 4D diffusion model. Another stream is optimization-based methods \cite{singer2023text,zeng2024stag4d,yin20234dgen,wu2024sc4d}. These methods generate 4D content through distilling spatial prior knowledge from pre-trained multi-view \cite{liu2023zero,shi2023zero123++} diffusion models. However, \textbf{\textit{previous methods only model temporal-spatial correlations using whole information in frames, but fail to explicitly differentiate between the dynamic and static regions within a frame}}. If static regions account for a significant portion, previous methods overlook dynamic information. Thus they tend to overfit static regions, resulting in a diminished capacity to perceive texture variations in dynamic regions. As shown in Fig.\ref{highlight} (a), previous work (STAG4D \cite{zeng2024stag4d}) produces results with blurry texture (e.g., wrinkles in clothes), particularly when confronted with B-type input video. 

To tackle above problems, we present a novel approach DS4D, which decouples dynamic and static features along temporal and spatial axes to enhance dynamic representations for high-quality 4D generation. In particular, we propose dynamic-static feature decoupling module (DSFD) to obtain decoupled features. We assume some regions of current frame features exhibit significant differences relative to reference frame features. Such regions always contain important knowledge of texture, shape variations, and motion trends under current timestamps. Therefore, it can be regarded as the dynamic feature of the current frame. In contrast, the remaining parts are the static features. Based on this assumption, we decompose the dynamic components between each frame feature and reference frame features along the temporal axes. The dynamic component (also termed dynamic features), as difference features between two frame features, is able to represent the dynamic information (as shown in Fig.\ref{highlight} (b)). At last, we acquire decoupled features driven by dynamic components and corresponding frame features. 

Note that spatial occlusion leads to the failure of dynamic components captured from a specific viewpoint to inadequately represent the inherent dynamic information in 4D space. To mitigate this issue, we design temporal-spatial similarity fusion module (TSSF) which engages in enhancing features' dynamic representations. Firstly, Gaussian points are initialized by utilizing a large reconstruction model \cite{xu2024instantmesh}. Then, point features are obtained by retrieving decoupled features for Gaussian points via view projection in TSSF. Subsequently, TSSF produces fused Gaussian features by adaptively selecting information on dynamic regions containing similar texture, shape, and motion representations from point feature space at same timestamps along spatial axes. Finally, fused Gaussian features with strong dynamic representation, are used for 4D generation.

The contributions can be summarized as follows:

\begin{itemize}
\item{To our best knowledge, we are the first to propose a framework DS4D that decouples dynamic-static information in frames along temporal-spatial axes to enhance dynamic representations for high-quality 4D generation.}
\item{Leveraging significant differences between frame features, we propose dynamic-static feature decoupling module (DSFD) to decouple dynamic and static features.}
\item{We tackle the issue of inadequate dynamic information in 4D space resulting from spatial occlusion by designing temporal-spatial similarity fusion module (TSSF). TSSF enhances dynamic representations in features.}

\item{Experimental results on Consistent4D dataset \cite{jiang2024consistentd} and Objaverse dataset \cite{deitke2023objaverse} demonstrate our DS4D outperforms other SOTA methods in terms of video quality, motion fidelity, and temporal-spatial consistency. Furthermore, the experiments on a complex real-world scenario dataset \cite{li2022neural} verify its effectiveness on 4D scenes.}
\end{itemize}

\section{Related Works}
\label{sec:related}

\begin{figure*}[!t]%
\centering
\includegraphics[width=1.0\textwidth]{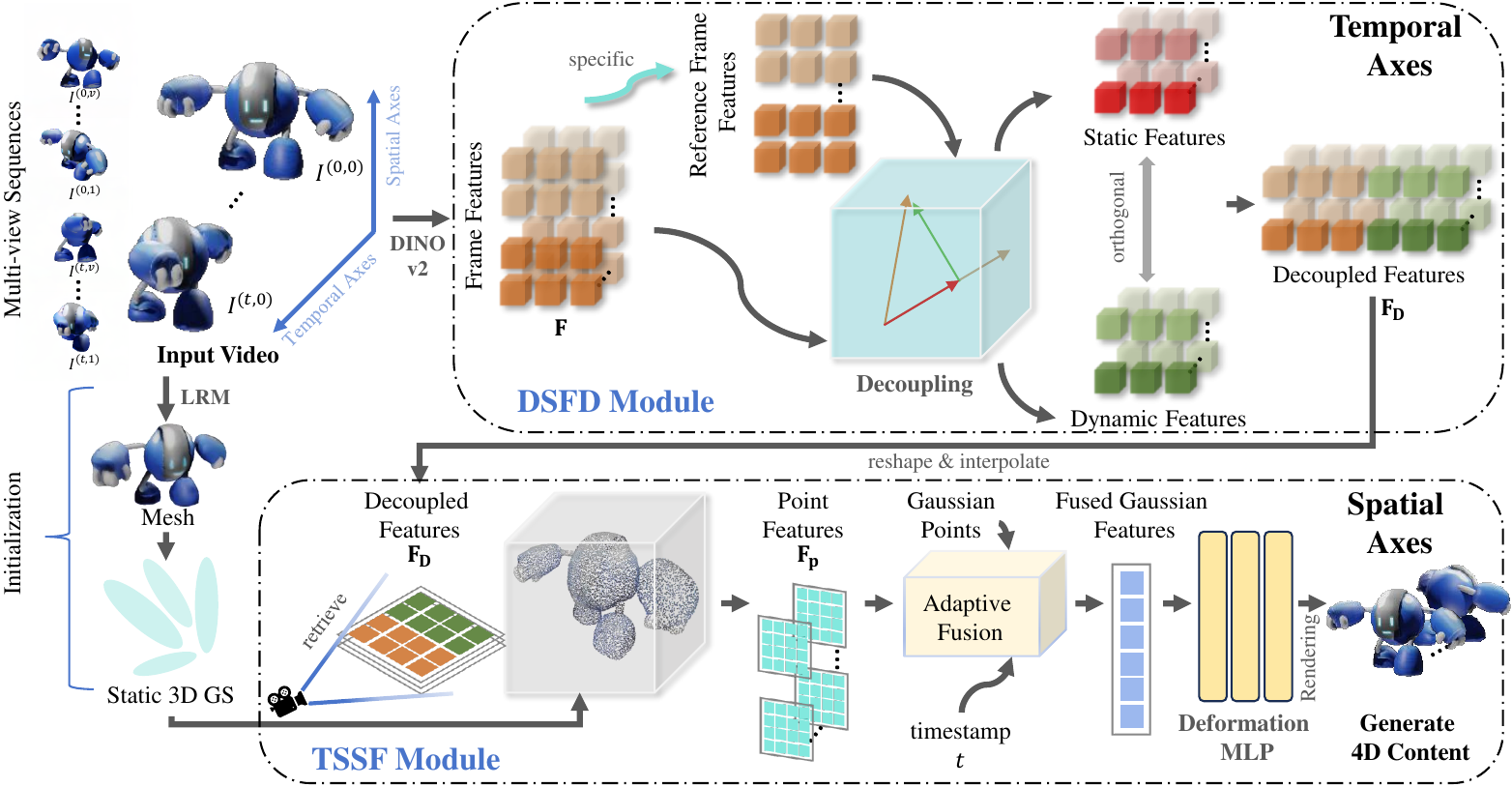}
\caption{\textbf{Overview of our proposed DS4D.} Given an input video and corresponding multi-view sequences, our DS4D decouples features of the frame at time $t$ based on the reference frame in DSFD module. Next, we acquire point features by retrieving each decoupled feature for Gaussian points via view projection, and we obtain fused Gaussian features by adaptively selecting similar dynamic information from point feature space in TSSF module. Finally, through Deformation MLP, our method generates 4D content. 
}\label{pipeline}
\end{figure*}

\subsection{3D Generation}

3D generation aims to produce 3D assets using images or text descriptions. The early works generate 3D objects in voxels \cite{choy20163d,xie2020pix2vox++,zhu2023umiformer,yang2023long}, meshes \cite{wang2018pixel2mesh,niemeyer2020differentiable,wen2022pixel2mesh++} or point clouds \cite{gadelha2018multiresolution,achlioptas2018learning,luo2021diffusion} forms. Recently, thanks to the popular application of NeRF \cite{mildenhall2021nerf,muller2022instant} and 3D Gaussian Splatting \cite{kerbl20233d,yu2024mip}, and the successful pre-trained text-to-image diffusion model \cite{saharia2022photorealistic} in 2D generation tasks, most works including DreamFusion \cite{poole2023dreamfusion} Dreamgaussian \cite{tang2024dreamgaussian} that focus on 3D generation try to use Score Distillation Sampling (SDS) loss \cite{poole2023dreamfusion} to explore the possibility of inspiring potential 3D-aware from diffusion. However, these works always suffer from over-saturation, over-smoothing, and multi-face problems. Because it is a challenge to distill the invisible views while ensuring multi-view consistency based on 2D diffusion. In contrast, Zero-1-to-3 \cite{liu2023zero} and Zero123++ \cite{shi2023zero123++} directly train a 3D-aware diffusion model using multi-view images and corresponding camera poses. Zero-1-to-3 and Zero123++ usually generate more stable and high-quality novel views thanks to the spatial perception built into them. Therefore, in our method, we use Zero123++ to predict the novel views at all timestamps to provide additional multi-view sequences for optimization. 

\subsection{4D Generation}

Compared to 3D generation, 4D generation is a more challenging task. Current 4D representations are two main streams, including NeRF-based \cite{mildenhall2021nerf,park2021nerfies,pumarola2021d,tretschk2021non,wu2022d,fang2022fast} and 3D Gaussians-based \cite{kerbl20233d,li2024spacetime,liang2023gaufre,luiten2024dynamic,li2024dreammesh4d,wu20244d,yang2023real}. Consistent4D \cite{jiang2024consistentd}, as the NeRF-based method, leverages the prior knowledge from pre-trained 2D diffusion models to optimize dynamic NeRF by Score Distillation Sampling (SDS) optimization. Dreamgaussian4D \cite{ren2023dreamgaussian4d}, STAG4D \cite{zeng2024stag4d} and SC4D \cite{wu2024sc4d} both attempt to introduce dynamic 3D Gaussians in 4D generation. Dreamgaussian4D notably reduces the cost of optimization time by improving the training strategies. STAG4D can generate anchor multi-view videos for optimization via a training-free strategy. SC4D optimizes sparse-controlled dynamic 3D Gaussians through SDS loss. However, these methods model temporal-spatial correlations using whole video but ignore the dynamic information in frames when static regions account for a large portion. It leads to them easily overfitting on static parts.

\section{Methodology}
\subsection{Overview}
In this section, we provide an overview of DS4D, which includes initialization in Sec.\ref{initial}, dynamic-static feature decoupling (DSFD) in Sec.\ref{dsfd}, and temporal-spatial similarity fusion (TSSF) in Sec.\ref{tscf}. The overall framework of our proposed method is illustrated in Fig.\ref{pipeline}.

\subsection{Initialization} \label{initial}

Formally, we begin with a single-view video. For each frame in the video, we adopt Zero123++ \cite{shi2023zero123++} to infer the $^{\dag}$pseudo multi-view images of frames. Then we obtain frame sequences, $\textbf{I}=\{I^{(0,0)}, I^{(0,1)},\dots,I^{(i,j)},\dots,I^{(t,v)}\}$, where $t$ represents the number of frames, and $v$ represents the number of views of each frame.

\textbf{Frame Feature Extraction.} Owing to the superior feature extraction capability of the visual foundation model, we employ it as the feature extractor to obtain high-quality features. Thanks to the strong feature extraction ability of DINOv2 \cite{oquab2023dinov2}, we leverage DINOv2 to extract frame features $\textbf{F}=\{f^{(0,0)}, f^{(0,1)},\dots,f^{(i,j)},\dots,f^{(t,v)}\}$ from sequences $\textbf{I}=\{I^{(0,0)}, I^{(0,1)},\dots,I^{(i,j)},\dots,I^{(t,v)}\}$, where $f^{(i,j)}\in \R^{P\times D}$, $P$ and $D$ are the number of tokens and token dimension, respectively. Each feature $f^{(i,j)}$ encodes the geometry and texture of the corresponding frame.

\renewcommand{\thefootnote}{\fnsymbol{footnote}}
\footnotetext[2]{The discussion about the impact of multi-view sequence quality and the robustness of our methods can be seen in supplementary material (Sec.~\ref{robust}).}

\textbf{Gaussian Points Initialization.} Previous works~\cite{zeng2024stag4d,wu2024sc4d} initialize Gaussian points randomly, leading to unstable topology during optimization and ultimately compromising the quality of results. To solve the above problem, we initialize static 3D Gaussian points through pre-defined point clouds. Specifically, we employ a large reconstruction model \cite{xu2024instantmesh} to generate the point clouds from the middle frame. This initialization approach provides a geometric prior and ensures the stability of subsequent optimization.

\subsection{Dynamic-Static Feature Decoupling (DSFD)} \label{dsfd} 
We employ a set of frame features $\textbf{F}$, which consists of all frame features and their corresponding multi-view features, to decouple dynamic-static features. Specifically, our decoupling process involves two aspects: 1) Finding out reference frame features; 2) Decoupling dynamic and static features along temporal axes.

\textbf{Finding out reference frame features.} Decoupling frame-by-frame features usually consumes considerable computation time. Hence, we specify special frames as references for decoupling. It is critical to determine whether reference frame features can accurately represent the semantic content of the entire video. Considering that we initialize static Gaussian points at the middle frame, the whole model needs to capture the texture, shape, and motion transformation from the middle frame to any frame during training and inference. Therefore, we choose the frame features $f^{(\frac{t}{2},j)}$ of the middle timestamp as one of the reference frame features, which represent the semantic. Furthermore, to represent the motion trends in video, we need a kind of feature that can describe the motion variations of the entire video. Hence, we also specify the $^{\dag}$average frame features $\bar{f}^{(\bar{t},j)}$ as reference frame features, which represent the average motion variations.

\textbf{Decoupling dynamic and static features along temporal axes.} To reliably decouple dynamic and static information in frame features, we need to obtain the regions of current frame features that exhibit significant differences relative to reference frame features. These regions contain important knowledge of texture, shape variations, and motion trends under current timestamps. The regions with significant differences mean dynamic parts between the two frame features, while regions with similar information mean static parts between the two frame features. Therefore, we propose the dynamic-static feature decoupling module (DSFD), as shown in Fig.\ref{decouple}.

Concretely, given specific $j$-th view in timestamp $i$, we project frame features $f^{(i,j)}$ onto reference frame features $r^{j}$. \textbf{The projection represents the semantic overlapping part of the two frames. In other words, regions in the two frames with no texture and shape transformations are identified as the static part.} The process is mathematically formulated as follows:

\begin{equation}
  \label{eq1}
  f_{\text{static}}^{(i,j)}=(\frac{f^{(i,j)}\cdot r^{j}}{\Vert r^{j}\Vert_2})\frac{r^{j}}{\Vert r^{j}\Vert_2}, 
\end{equation} 

\begin{figure}[t]%
\centering
\includegraphics[width=0.4\textwidth]{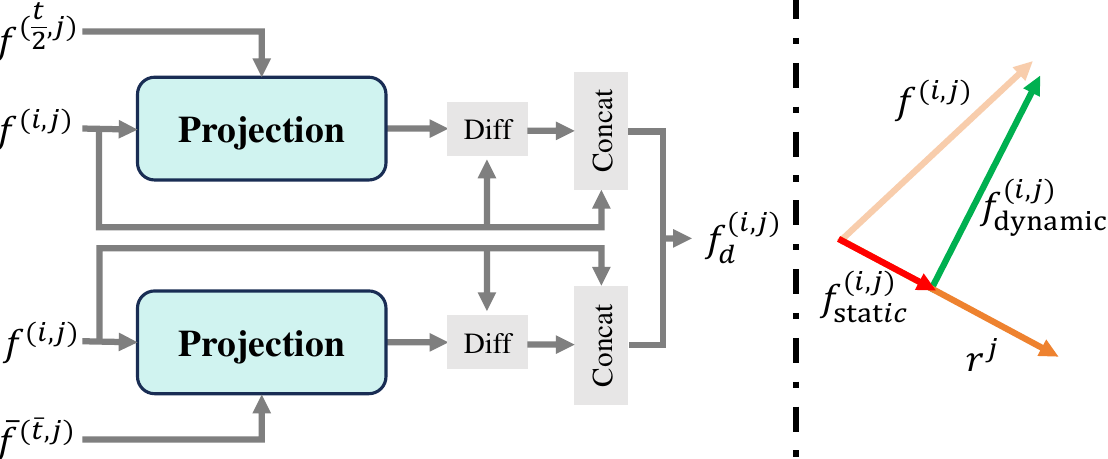}
\caption{An overview of the decoupling architecture in DSFD. The decoupling architecture is shown on the left. The demonstration of projection is shown on the right.}\label{decouple}
\end{figure}

where $\cdot$ denotes dot product.

The difference between current frame features and static features means the orthogonal vector. Then, we identify the orthogonal vector as the semantic difference between the current frame and the reference frame. Hence, the orthogonal vector can be regarded as dynamic features:

\begin{equation}
  \label{eq2}
  f_{\text{dynamic}}^{(i,j)}=f^{(i,j)}-f_{\text{static}}^{(i,j)}. 
\end{equation}

At each timestamp, we obtain the dynamic features in a particular view. Dynamic features contain significant difference information, which represents the motion knowledge of the current frame relative to the reference frame. Meanwhile, frame features contain texture and shape attributes of the current frame, which are useful for subsequently texture and shape learning. Hence, we leverage dynamic features to supplement dynamic information in the frame features while retaining the inherent attributes of the frame features. 

Here, we append each token of the current frame features $f^{(i,j)}$ with dynamic features $f_{\text{dynamic}}^{(i,j)}$. Then, the appended features $f_d^{(i,j)}$ are integrated based on both the temporal $i$ and spatial $j$, which are named decoupled features $\textbf{F}_{\textbf{D}}=\{f_d^{(0,0)},f_d^{(0,0)},\dots,f_d^{(i,j)},\dots,f_d^{(t,v)}\}$, where $t$ and $v$ denote the number of frames and number of views, respectively.

\subsection{Temporal-Spatial Similarity Fusion (TSSF)} \label{tscf} 
Since the features in $\textbf{F}_{\textbf{D}}$ obtained from a particular viewpoint do not adequately represent the complete dynamic information in the 4D space, we design temporal-spatial similarity fusion module (TSSF) to adaptively select similar dynamic information from decoupled features under different views at the same timestamp.

\renewcommand{\thefootnote}{\fnsymbol{footnote}}
\footnotetext[2]{Average frame features $\bar{f}^{(\bar{t},j)}$ are derived from the average of all frame features from a particular view.}

Inspired by \cite{melas2023pc2}, we retrieve the decoupled features from Gaussian points by view projection. Then we integrate decoupled features into point features based on the given camera pose. The Gaussian points are well-aligned with decoupled features thanks to the projection operations. Therefore, point features $\textbf{F}_{\textbf{p}}=\{f_p^{(0,0)},f_p^{(0,1)},\dots,f_p^{(i,j)},\dots,f_p^{(t,v)}\}$ encapsulate extensive dynamic information from decoupled features. In fact, among different viewpoints, the texture, shape, and motion representations from the same spatial area of the object are similar in the same timestamps. Based on this fact, utilizing the semantic similarities in point features across viewpoints and aggregating them contributes to the enhancement of dynamic representations in point features. Hence, we propose an adaptive fusion in Fig.\ref{fusion}.

\begin{figure}[t]%
\centering
\includegraphics[width=0.4\textwidth]{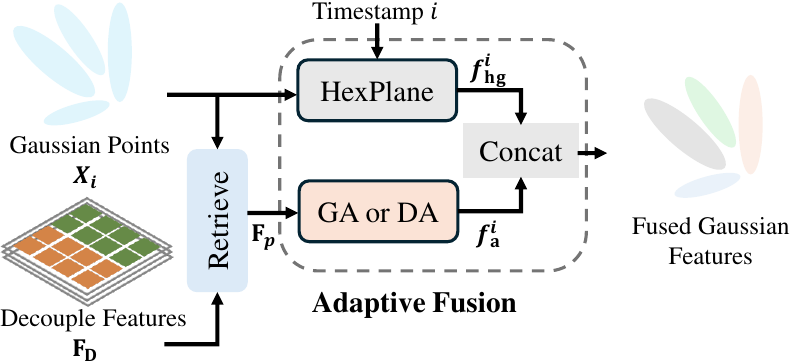}
\caption{An overview of adaptive fusion in TSSF.}\label{fusion}
\end{figure}

At the same timestamp, we design global awareness fusion (GA) in adaptive fusion to merge similar dynamic information from point features along spatial axes. However, under conditions of significant differences among multi-views, the effect of merging in GA is affected by that dynamic regions are noticeably obscured in specific viewpoints. Therefore, we need to reduce the impact of point features under these views for merging while retaining the useful dynamic information in these features. Based on this condition, we design distance awareness fusion (DA).

\textbf{Global awareness (GA).} At time $i$, the fully-connection layer (FC) generates score maps $\textbf{W}=\{w^{(0,0)},\dots,w^{(t,v)}\}$ for the point features of each view,

\begin{equation}
  \label{equ:fc1}
  \textbf{W}=\mathtt{Softmax}(\mathtt{FC}(\textbf{F}_{\textbf{p}})). 
\end{equation} 

Then, we fuse them into fused point features by the weighted summation of all point features according to their score map. The fused features $f_a^i$ can be calculated as:
\begin{equation}
  \label{equ:fc2}
  f_a^i=\sum_{j=0}^{v}{w^{(i,j)}f_p^{(i,j)}}. 
\end{equation} 

\textbf{Distance awareness (DA).} Frames from single-view video are regarded as front view. The front frame reveals real and visible motion areas of the object at various timestamps compared to the other views. This information within the front frame requires preservation as much as possible while supplementing the useful dynamic information from other views for accurancy motion prediction. Thus, at time $i$, we calculate the $L_1$ distance between point features from the front view and point features from the other views. The distance quantifies the difference information between point features extracted from the front view and those derived from other viewpoints. The difference information contains dynamic information that cannot be represented by the point features of the front view. Then, we append point features from other views with distance features and merge them by \cref{equ:fc1} and \cref{equ:fc2}. The fused point features from other views are appended to point features from the front view. We merge these features by \cref{equ:fc1} and \cref{equ:fc2} to acquire the fused point features $f_a^i$. Through the above pre-processing, we reduce the impact of point features under other views according to the front view while retaining useful dynamic information in point features under other views.


In addition to fused features through adaptive fusion, we also present dynamic Gaussian features as 3D Gaussian features with HexPlane \cite{cao2023hexplane} for inherently regularizing features of fields and guaranteeing their smoothness. Formally, the dynamic Gaussian features at time $i$ can be defined as:

\begin{equation}
  \label{equ:sig1}
  f_{\text{hg}}^{i}=\mathtt{HexPlane}(X_i,s_i,\gamma_i,\sigma,\zeta), 
\end{equation} 
where $X_i=(x_i,y_i,z_i)$ denotes the position of Gaussian points, $s_i$ and $\gamma_i$ denote scale and rotation at time $i$. $\sigma$ and $\zeta$ represent opacity and spherical harmonic coefficients of the radiance. Then, we combine dynamic Gaussian features $f_{\text{hg}}^{i}$ and fused point features $f_a^i$ at time $i$, and map them to fused Gaussian features via learnable linear transformation. The fused Gaussian features represent Gaussian intrinsic properties while capturing rich dynamic information. Therefore, fused Gaussian features are essential for accurately predicting deformation using Deformation MLP.

\subsection{Training Objectives} Following \cite{zeng2024stag4d}, we calculate SDS loss and the photometric loss based on rendered views and ground truth images. Inspired by \cite{hong2023lrm,zou2024triplane,xu2024instantmesh}, we introduce the LPIPS loss to minimize the similarity between pseudo multi-view images and rendered views. More details can be found in the supplementary material.

\section{Experiments}

\begin{table*}[t]
	\centering
	\renewcommand\arraystretch{1.}
	\scalebox{0.8}{
	\begin{tabular}{c|c|cccc|cccc}
        \Xhline{1.5px}
		\multirow{2}{*}{\textbf{Methods}} & \multirow{2}{*}{\textbf{Optimization}} & \multicolumn{4}{c|}{\textbf{Consistent4D dataset}} & \multicolumn{4}{c}{\textbf{Objaverse dataset}} \\ \cline{3-10}
          & &\multicolumn{1}{c}{\textbf{CLIP $\uparrow$}} & \multicolumn{1}{c}{\textbf{LPIPS $\downarrow$}} & \multicolumn{1}{c}{\textbf{FVD $\downarrow$}} & \multicolumn{1}{c|}{\textbf{FID-VID $\downarrow$}} & \multicolumn{1}{c}{\textbf{CLIP $\uparrow$}} & \multicolumn{1}{c}{\textbf{LPIPS $\downarrow$}} & \multicolumn{1}{c}{\textbf{FVD $\downarrow$}} & \multicolumn{1}{c}{\textbf{FID-VID $\downarrow$}}
         \\ \hline
        \textbf{Consistent4D \cite{jiang2024consistentd}} & \Checkmark & 0.9085 & 0.1316 & 1041.2242 & 28.6471 & 0.8491 & 0.2222 & 1814.5652 & 48.7921 \\
        \textbf{Dreamgaussian4D \cite{ren2023dreamgaussian4d}} & \Checkmark & 0.9145 & 0.1517 & 844.9087 & 37.9977 & 0.8127 & 0.2017 & 1545.3009 & 58.3686  \\
        \textbf{STAG4D \cite{zeng2024stag4d}} & \Checkmark & 0.9078 & 0.1354 & 986.8271 & 26.3705 &  0.8790 & 0.1811 & 1061.3582 & 30.1359  \\
        \textbf{SC4D \cite{wu2024sc4d}} & \Checkmark & 0.9117 & 0.1370 & 852.9816 & 26.4779 & 0.8490 & 0.1852 & 1067.7582 & 40.5130 \\
        \textbf{4Diffusion \cite{zhang20244diffusion}} & \XSolidBrush & 0.8734 & 0.2284 & 1551.6363 & 149.6170 & - & - & - & - \\ 
        \textbf{L4GM \cite{ren2024l4gm}} & \XSolidBrush & 0.9158 & 0.1497 & 898.0604 & 31.4996 & - & - & - & - \\        
        \hline
        \textbf{DS4D-GA (Ours)} & \Checkmark & 0.9206 & 0.1311 & 799.9367 & 26.1794 & 0.8868 & 0.1761 & 890.2646 & 26.6717 \\
        \textbf{DS4D-DA (Ours)} & \Checkmark & \textbf{0.9225} & \textbf{0.1309} & \textbf{784.0235} & \textbf{24.0492} & \textbf{0.8881} & \textbf{0.1759} & \textbf{870.9489} & \textbf{25.3836} 
	    \\ \Xhline{1.5px}
	\end{tabular}}
	\caption{Evaluation and comparison of the performance on Consistent4D dataset and Objaverse dataset. The best score is highlighted in bold. All the experiments of the methods are carried out using the code from their official GitHub repository. For a fair comparison, the experiment of and L4GM on Objaverse dataset are disregarded since they are inference-based methods trained on this dataset.}
\label{total_result}
\end{table*}

\begin{figure*}[t]
    \centering
	\begin{minipage}{1\linewidth}
        \centerline{\includegraphics[width=1.\linewidth]{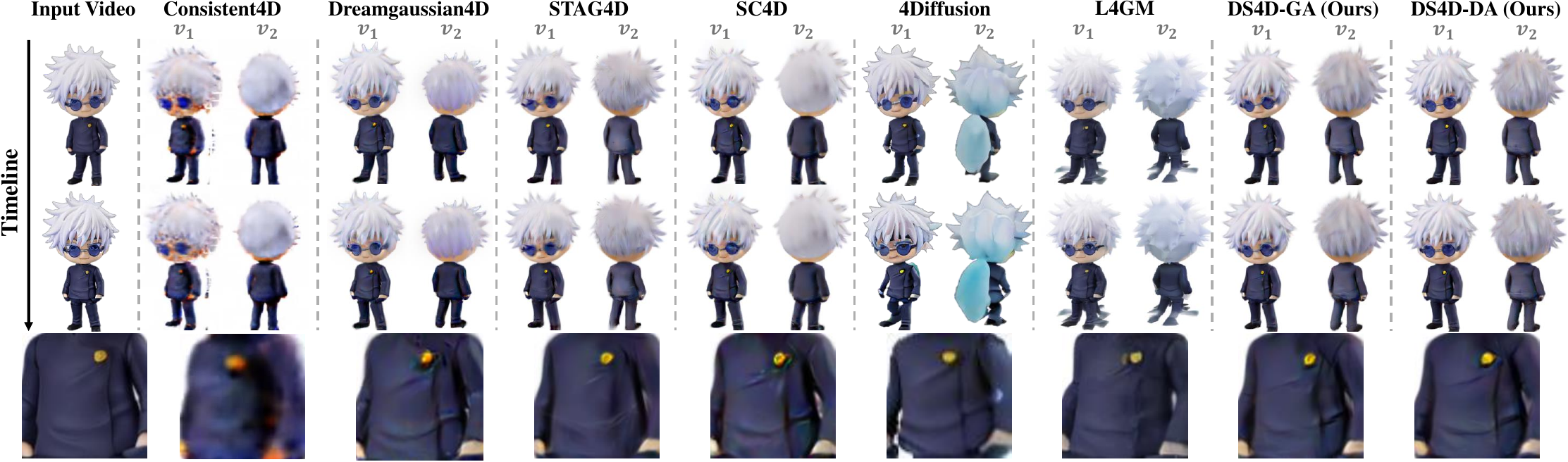}}
	\end{minipage}
	\\ \vspace{0.5mm}
	\begin{minipage}{1\linewidth}
        \centerline{\includegraphics[width=1.\linewidth]{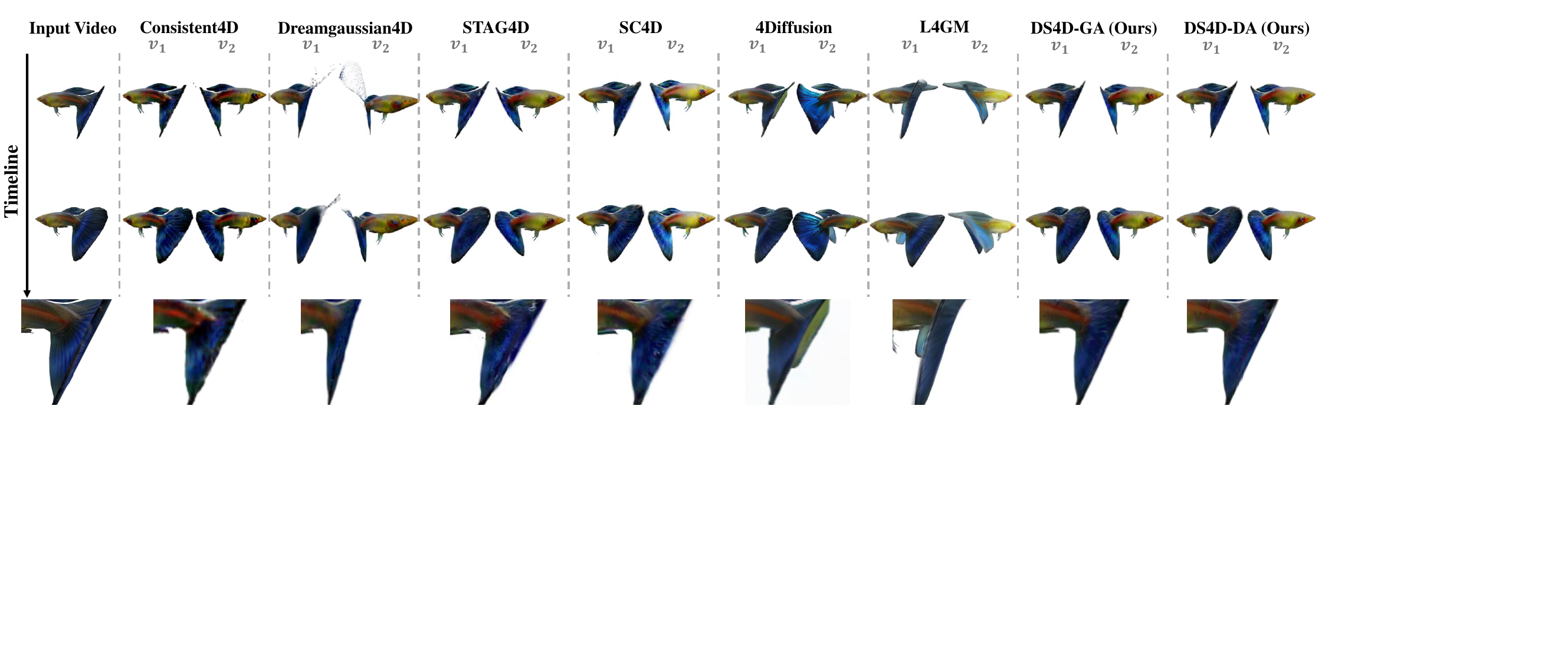}}
	\end{minipage}
	\caption{Qualitative comparison on video-to-4D generation. For each method, we render results under two novel views at two timestamps. \textbf{More comparison examples, and some of the 360$^\circ$ generation results (video file) can be found in supplementary materials.}}
\label{show_results}
\end{figure*}

\subsection{Implementation Details}
The implementation details are provided in the section. 

\textbf{Datasets} In our article, we use \textbf{four} challenging datasets. 1) We utilize the dataset provided by Consistent4D. 2) We introduce a subset of Objaverse \cite{deitke2023objaverse,liang2024diffusion4d}. The objects in Objaverse have more complexity motion than Consistent4D dataset. 3) We introduce some challenging videos from online sources. 4) In Sec.\ref{real-world}, to evaluate the effectiveness of our method in real-world scenarios, we utilize three real-world scenarios provided by Neu3D's \cite{li2022neural} dataset. More details can be found in supplementary material.

\textbf{Metrics} Following Consistent4D \cite{jiang2024consistentd} and STAG4D \cite{zeng2024stag4d}, we measure the 4D generation quality of our methods using CLIP, LPIPS, FVD and FID-VID. In Sec.\ref{real-world}, following \cite{wu20244d}, we evaluate our results using PSNR, LPIPS, SSIM, and structural dissimilarity index measure (D-SSIM).

\textbf{Training} Our two models, including DS4D-GA (using GA in TSSF) and DS4D-DA (using DA in TSSF) use the same training setting. The training details can be found in Supplementary Material.

\subsection{Comparisons with Existing Methods}
In this section, we compare our methods DS4D-GA and DS4D-DA with the SOTA methods on Consistent4D dataset and Objaverse dataset, including Consistent4D \cite{jiang2024consistentd}, Dreamgaussian4D \cite{ren2023dreamgaussian4d}, STAG4D \cite{zeng2024stag4d}, SC4D \cite{wu2024sc4d}, 4Diffusion \cite{zhang20244diffusion} and L4GM \cite{ren2024l4gm}. We conduct qualitative and quantitative comparisons, respectively. The superior performance demonstrates the effectiveness of our methods.

\textbf{Quantitative comparisons.} The quantitative results are shown in Tab.~\ref{total_result}. Our methods DS4D-GA and DS4D-DA, consistently outperform other methods in all metrics. Specifically, our methods notably exceed the SOTAs in FVD, indicating that our generation results have fewer temporal artifacts than others. Furthermore, the experiments on Objaverse dataset present that our methods significantly outperform other SOTAs by a large margin. It demonstrates our methods are superior in terms of temporal-spatial consistency, fidelity, and quality of generation results and underscores the robustness of our methods. In summary, such significant improvements are attributed to our method of decoupling dynamic-static features, which can explicitly distinguish dynamic and static regions within frame features.

\begin{figure}[h]%
\centering
\includegraphics[width=0.46\textwidth]{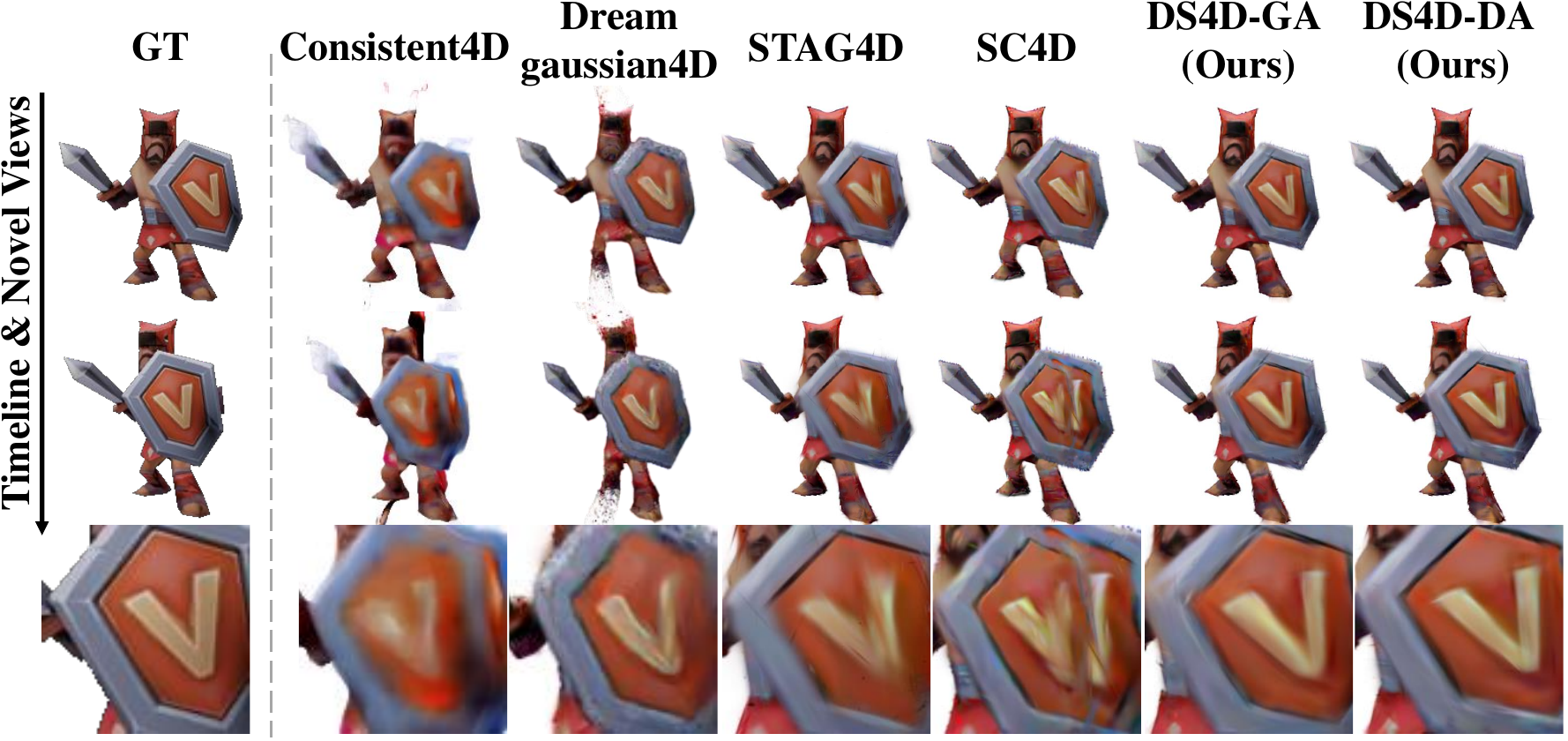}
\caption{Qualitative comparison on video-to-4D generation based on Objaverse dataset.}\label{objaverse1}
\end{figure}

\textbf{Qualitative comparisons.} The qualitative results on Consistent4D dataset are presented in Fig.~\ref{show_results}. Besides, the results on Objaverse dataset are shown in Fig.\ref{objaverse1}. Obviously, whether in the example with a larger proportion of static regions (the person in Fig.~\ref{show_results} and warrior in Fig.\ref{objaverse1}) or the example with a balanced proportion of dynamic and static regions (guppie in Fig.~\ref{show_results}), the results, generated by Consistent4D, Dreamgaussian4D, STAG4D and SC4D, have different degrees of blurriness in the details, especially in the areas with motion trends in the current and subsequent frames. Although 4Diffusion and L4GM generate videos with clearer textures than other previous methods with the help of a large amount of 4D data prior, it is easy to generate the abnormal shape or textures with inconsistent details in some areas (e.g, the back of the person). These methods do not differentiate between dynamic and static information, leading them to easily overlook information in dynamic regions when faced with large proportions of static. In contrast, our methods address the challenges arising from varying proportions of dynamic and static regions in frames, resulting in the achievement of high-quality 4D generation.

\begin{table*}[t]
	\centering
	\renewcommand\arraystretch{1.}
	\scalebox{0.75}{
	\begin{tabular}{c|c|cccc|cccc}
        \Xhline{1.5px}
		\multirow{2}{*}{\textbf{Methods}} & \multirow{2}{*}{\textbf{Point Initialization}} & \multicolumn{4}{c|}{\textbf{Consistent4D dataset}} & \multicolumn{4}{c}{\textbf{Objaverse dataset}} \\ \cline{3-10}
          & &\multicolumn{1}{c}{\textbf{CLIP $\uparrow$}} & \multicolumn{1}{c}{\textbf{LPIPS $\downarrow$}} & \multicolumn{1}{c}{\textbf{FVD $\downarrow$}} & \multicolumn{1}{c|}{\textbf{FID-VID $\downarrow$}} & \multicolumn{1}{c}{\textbf{CLIP $\uparrow$}} & \multicolumn{1}{c}{\textbf{LPIPS $\downarrow$}} & \multicolumn{1}{c}{\textbf{FVD $\downarrow$}} & \multicolumn{1}{c}{\textbf{FID-VID $\downarrow$}}
         \\ \hline
        \textbf{A} & \XSolidBrush & 0.9133 & 0.1341 & 953.6300 & 27.3747 & 0.8736 & 0.1816 & 1072.1292 & 28.2410 \\
        \textbf{B} & \Checkmark & 0.9151 & 0.1313 & 913.371 & 27.1357 & 0.8763 & 0.1801 & 1062.9398 & 28.0977 \\
        \textbf{C. w/ LPIPS Loss} & \Checkmark & 0.9163 & 0.1311 & 899.5714 & 27.0836 & 0.8773 & 0.1804 & 1016.2576 & 27.8062 \\
        \textbf{D. w/ Frame Features} & \Checkmark & 0.9174 & 0.1350 & 888.6579 & 26.8486 & 0.8805 & 0.1778 & 1005.7503 & 27.6334 \\
        \textbf{E. w/ DSFD} & \Checkmark & 0.9186 & 0.1333 & 861.6075 & 26.5403 & 0.8827 & 0.1765 & 989.0834 & 26.9199 \\
        \hline
        \textbf{F. w/ TSSF-Average Pooling} & \Checkmark & 0.9194 & 0.1313 & 839.6600 & 26.5071 & 0.8848 & 0.1761 & 951.9127 & 26.8412\\ 
        \textbf{G. w/ TSSF-GA (Ours)} & \Checkmark & 0.9206 & 0.1311 & 799.9367 & 26.1794 & 0.8868 & 0.1761 & 890.2646 & 26.6717 \\
        \textbf{H. w/ TSSF-DA (Ours)} & \Checkmark & \textbf{0.9225} & \textbf{0.1309} & \textbf{784.0235} & \textbf{24.0492} & \textbf{0.8881} & \textbf{0.1759} & \textbf{870.9489} & \textbf{25.3836} 
	    \\ \Xhline{1.5px}
	\end{tabular}}
	\caption{The ablation experiments on Consistent4D dataset and Objaverse dataset. Each setup is based on a modification of the immediately preceding setups. The best score is highlighted in bold.}
\label{ablation_result}
\end{table*}

\subsection{Ablation Experiments}
To evaluate the effectiveness of different components in our methods, we conduct ablation experiments on Consistent4D and Objaverse datasets as shown in Tab.\ref{ablation_result}. More analysis can be found in supplementary material.

\begin{figure}[t]%
\centering
\includegraphics[width=0.45\textwidth]{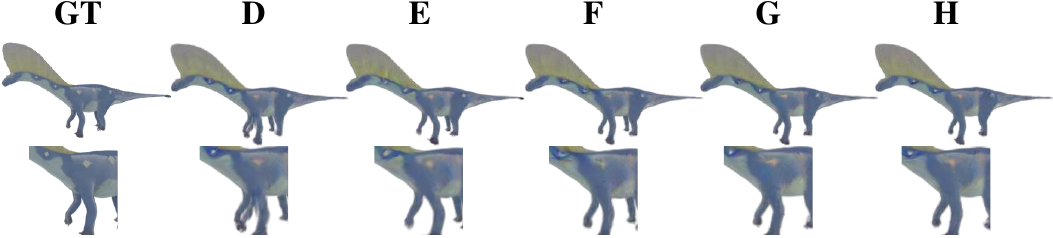}
\caption{Ablation on different experiment settings from Tab.\ref{ablation_result}.}\label{ablation}
\end{figure}

\begin{figure}[t]%
\centering
\includegraphics[width=0.45\textwidth]{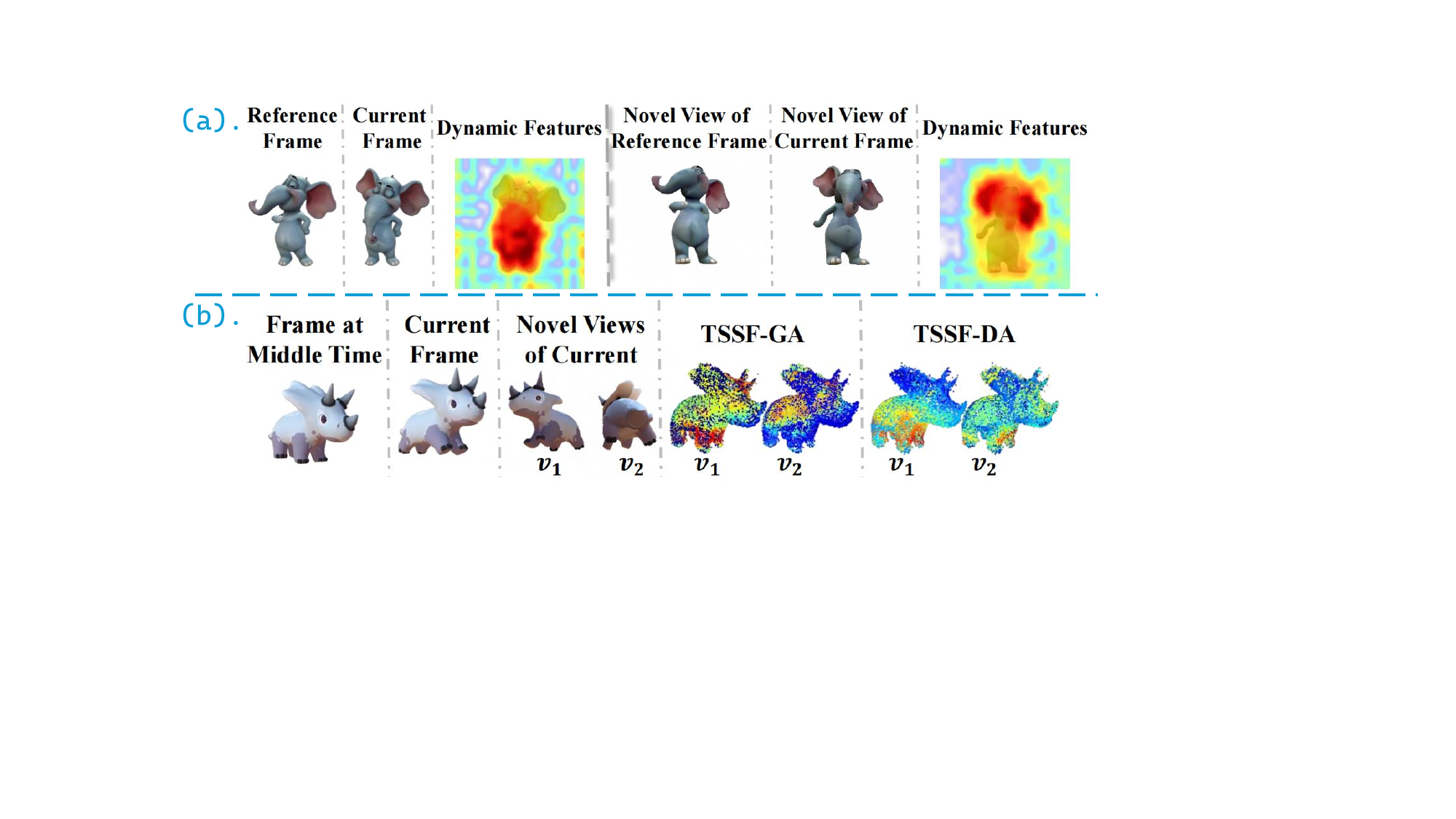}
\caption{(a) Visualization on the heatmap of dynamic features in DSFD. The red region highlights the primary region of interest in dynamic features. (b) The score map of point features in TSSF-GA and TSSF-DA. The red area indicates high attention to dynamic information in point features of a specific view.}\label{visualization}
\end{figure}

\textbf{Effect of DSFD and Decoupling.} Validating the effect of DSFD also means Validating the effect of decoupling since DSFD consists of decoupling. Therefore, we conduct experiments in two aspects: a) Adding frame features in the model without decoupling (labeled as D). b) Performing decoupling on frame features in the model (labeled as E, also termed DSFD). In particular, the performance of D and E both have significant improvement compared to A and B, especially FVD score. For D, the frame features provide enough structure prior knowledge and texture information for Deformation MLP. It assists Deformation MLP in capturing shape deformation. Nevertheless, D fails to differentiate between dynamic and static regions of frames clearly. Consequently, D is prone to overfitting in static areas, resulting in blurry details. For example, the blurred details of legs from D in Fig.\ref{ablation}. In contrast, DSFD, which decouples dynamic-static features, mitigates the issue.

\textbf{Effect of TSSF.} To validate the effect of TSSF, we add the TSSF based on model E, including model F, G (TSSF-GA) and H (TSSF-DA). Among them, we replace adaptive fusion with average pooling in model F. Specifically, we observe that E, G, H can generate more texture details than other models. Because they mitigate the problem brought by spatial occlusion by aggregating dynamic components from different viewpoints. However, F only averages the point features from DSFD along spatial axes, which ignores merging similar dynamic information. The model G and F adaptively select dynamic information, which plays a crucial role in enhancing dynamic representations. Compared with E, the texture of 4D content generated by F and H is more refined in Fig.\ref{ablation}. This underscores the efficacy of merging similar dynamic information in TSSF, including TSSF-GA and TSSF-DA.

\subsection{Visualization} 
In this section, we perform several visualizations of features in DSFD and TSSF, respectively.

\textbf{DSFD.} Fig.\ref{visualization} (a) presents the heatmap of dynamic features obtained by DSFD decoupling features from the current and reference frame features. The red area indicates the primary region of interest in the features. We observe the motion trends come from the trunk and body of elephant in the front view (left of the figure). Thus, dynamic features show strong concern in this area. Limited to the viewpoint range, the motion trajectory for novel views (right of the figure) is more interested in the elephant's head. No doubt, dynamic features also present a high response in similar areas. In conclusion, our method can acquire accurate dynamic features using DSFD, as supported by the visualization results.

\textbf{TSSF.} Fig.\ref{visualization} (b) shows the score map of adaptively selecting similarity dynamic information from point features of different views by TSSF-GA and TSSF-DA.  Specifically, the leg movements of the triceratops indicate the primary trend in motion between the middle and current time. The red area indicates the high attention to dynamic information in point features of a specific view. Meanwhile, this area is also similar to the dynamic area of other views in Fig.\ref{visualization} (b). Hence, it reveals that the two approaches can capture a certain degree of similar dynamic information from different viewpoints. However, TSSF-GA is interested in the back of triceratops rather than legs since the front legs are nearly invisible in the novel view $v_2$. The unseen dynamic region in the novel view influences the accuracy of selecting similar dynamic information. In contrast, TSSF-DA predicts highest scores on the legs under view $v_1$ and $v_2$, thanks to reducing the impact of novel views in TSSF-DA. It indicates TSSF-DA can alleviate issues caused by the fact that dynamic regions are noticeably obscured in novel views.

\begin{figure}[]%
\centering
\includegraphics[width=0.4\textwidth]{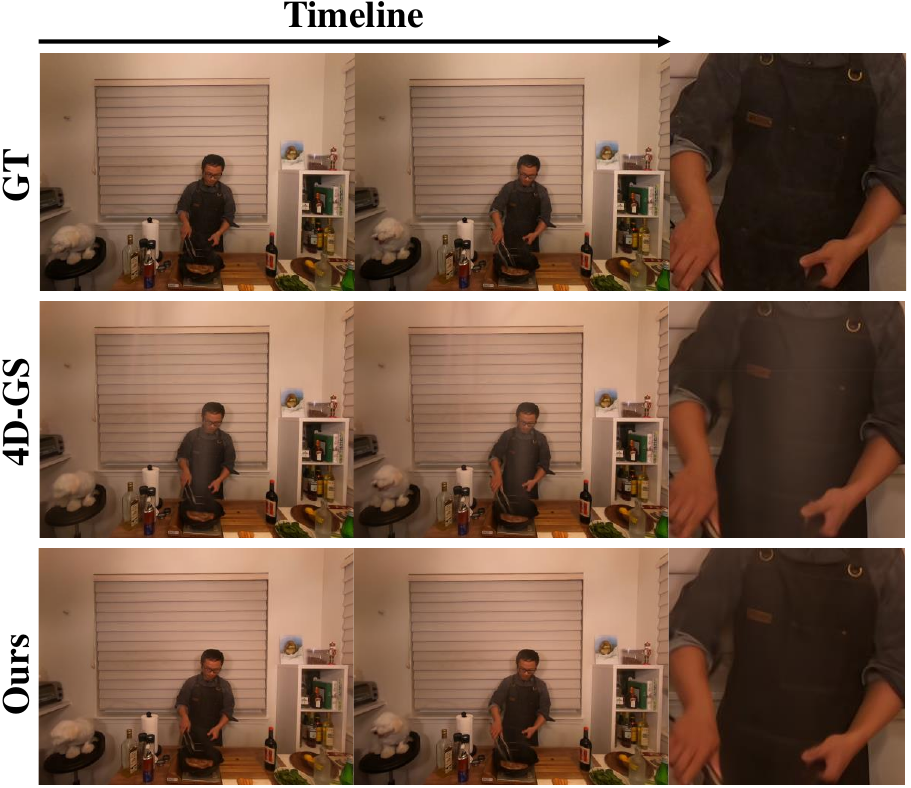}
\caption{Visualization of flame steak compared with 4D-GS.}\label{discussion2}
\end{figure}
\section{Discussion on Real-World Scenario} \label{real-world}
Generating a dynamic scene from input real-world videos is an essential task. In this section, we conduct experiments on three scenes from Neu3D’s dataset to demonstrate the effectiveness of DSFD and TSSF on complexity real-world dynamic scene. We set 4D-GS \cite{wu20244d} as the baseline, which has shown promising results in this task. To fair comparison, we directly insert our DSFD and TSSF into 4D-GS. More details can be found in supplementary material.

\begin{table}[]
	\centering
	\renewcommand\arraystretch{0.95}
	\scalebox{0.85}{
    \begin{tabular}{c|cccc}
	    \hline
		\multicolumn{1}{c|}{\textbf{\makecell{Method}}} &
		\multicolumn{1}{c}{\textbf{PSNR $\uparrow$}} & \multicolumn{1}{c}{\textbf{SSIM $\uparrow$}} & \multicolumn{1}{c}{\textbf{D-SSIM $\downarrow$}} & \multicolumn{1}{c}{\textbf{LPIPS $\downarrow$}} \\ \hline
		4D-GS \cite{wu20244d} & 32.1598 & 0.9483 & 0.0132 & \textbf{0.1422} \\
		Ours & \textbf{32.3964} & \textbf{0.9494} & \textbf{0.0123} & 0.1434 \\ \hline
	\end{tabular}}
	\caption{Evaluation of the performance on Neu3D's dataset.}
\label{discussion1}
\end{table}

The results are shown in Tab.\ref{discussion1} and Fig.\ref{discussion2}. Our method achieves competitive results, demonstrating significant potential even for intricate real-world scenarios. Due to static nature of a significant portion of the real-world scene, the previous method easily encounters challenges in exploring dynamic regions, consequently impacting the quality of outputs. Hence, it is necessary to decompose dynamic and static regions to mitigate this issue. In essence, significant differences exhibit texture variations and motion trends between frames. Thus it is beneficial to utilize this nature to decouple dynamic-static information at feature-level, which enables us to enhance fine-grained dynamic representations. Indeed, there are additional methods available to decouple effectively. For instance, we can introduce optical flow and utilize 3D-aware foundation model \cite{xu2024depthsplat,yang2024depthv2} to extract depth features used to decouple.

\section{Conclusion}
In this paper, we propose a novel framework DS4D, which decouples dynamic-static information along temporal-spatial axes to enhance dynamic representations for high-quality 4D generation. Its DSFD decouples features by regarding the regions of current features that possess significant differences relative to reference frame features as dynamic features. Moreover, TSSF is designed to enhance dynamic representations by selecting similar dynamic information. Overall, our method can produce high-quality 4D content and shows promise in 4D scene generation.

~\\
\setlength{\parindent}{0pt} { \textbf{Acknowledgements} This work is supported by the National Key Research and Development Plan under Grant 2021YFE0205700, Science and Technology Development Fund of Macau project 0070/2020/AMJ, 00123/2022/A3, 0096/2023/RIA2.
}
{\small
\bibliographystyle{unsrt}
\bibliography{main}
}

\clearpage
\setcounter{page}{1}
\maketitlesupplementary

\section{The 360$^\circ$ Results Generated by Our methods}

\textbf{We strongly recommend checking the video file in the supplementary material.} Some of the 360$^\circ$ results have been attached in the video of the supplementary material. In future official versions, we will provide more 360$^\circ$ results and comparisons.

\section{Training Objectives}
In this section, we present more details on training objectives. Please note that each loss in our loss function has a clear derivation from previous works. The theoretical derivation can be found in corresponding previous works, e.g., \cite{zhang2018unreasonable,tang2023dreamgaussian,zeng2024stag4d} etc. Hence, we will not provide a detailed derivation process in our manuscript.

\textbf{Score distillation sampling loss.} Following \cite{zeng2024stag4d}, we employ multi-view score distillation sampling (SDS) loss using rendered images under camera poses of pseudo multi-view images and input video.In detail, there are rendered multi-view images $I=\{I^{(i,1)},\dots,I^{(i,j)}\}$ under 6 camera poses, where $i$ denotes the timestamps and $j$ denotes the number of views. Formally, SDS loss can be defined as:

\begin{equation}
\begin{aligned}
      \mathcal{L}_{SDS} &=\alpha_1\mathcal{L}_{SDS}^{pseudo}+\alpha_2\mathcal{L}_{SDS}^{real}\\&=\alpha_1\mathcal{L}_{SDS}(\phi,I^{(i,j)})+\alpha_2\mathcal{L}_{SDS}(\phi,I_{real}^{i})  
\end{aligned}
\end{equation}

where $\alpha_1$ and $\alpha_2$ are hyperparameters, $I_{real}^{i}$ is the rendered image under camera pose of input video at time $i$. 

\textbf{Photometric loss.} Following \cite{tang2023dreamgaussian,zeng2024stag4d}, we compute the reconstruction loss $\mathcal{L}_{rec}$ between rendered images and pseudo multi-view images, and the foreground mask $\mathcal{L}_{mask}$.

\textbf{LPIPS loss.} We introduce the LPIPS loss $\mathcal{L}_{lpips}$ \cite{zhang2018unreasonable} to compute the feature similarity between pseudo multi-view images and corresponding rendered images. We leverage VGG \cite{simonyan2014very} as backbone to extract image features.

\textbf{Overall loss.} Based on above loss functions, we obtain the final overall loss $\mathcal{L}$:

\begin{equation}
    \mathcal{L} = \lambda_1\mathcal{L}_{SDS}+\lambda_2\mathcal{L}_{rec}+\lambda_3\mathcal{L}_{mask}+\lambda_4\mathcal{L}_{lpips}
\end{equation}

where $\lambda_1,\lambda_2,\lambda_3,\lambda_4$ are hyperparameters.

\section{Training Details of DS4D-GA and DS4D-DA}
We train our two models DS4D-GA (using GA in TSSF) and DS4D-DA (using DA in TSSF) under the same training setting. During the initial 1,000 iterations, we train our models except TSSF and deformation MLP. Subsequently, the models with TSSF and deformation MLP are optimized over 6,000 additional iterations. For the deformation MLP, we employ each MLP with 64 hidden layers and 32 hidden features. The learning rate of TSSF and deformation MLP is set to $1.6\times10^{-4}$ and is decayed to $1.6\times10^{-6}$. Following \cite{zeng2024stag4d}, the top $2.5\%$ of points are densified with the most accumulated gradient. The overall training process costs approximately 3 hours on a V100 GPU. 

\begin{figure}[t]%
\centering
\includegraphics[width=0.45\textwidth]{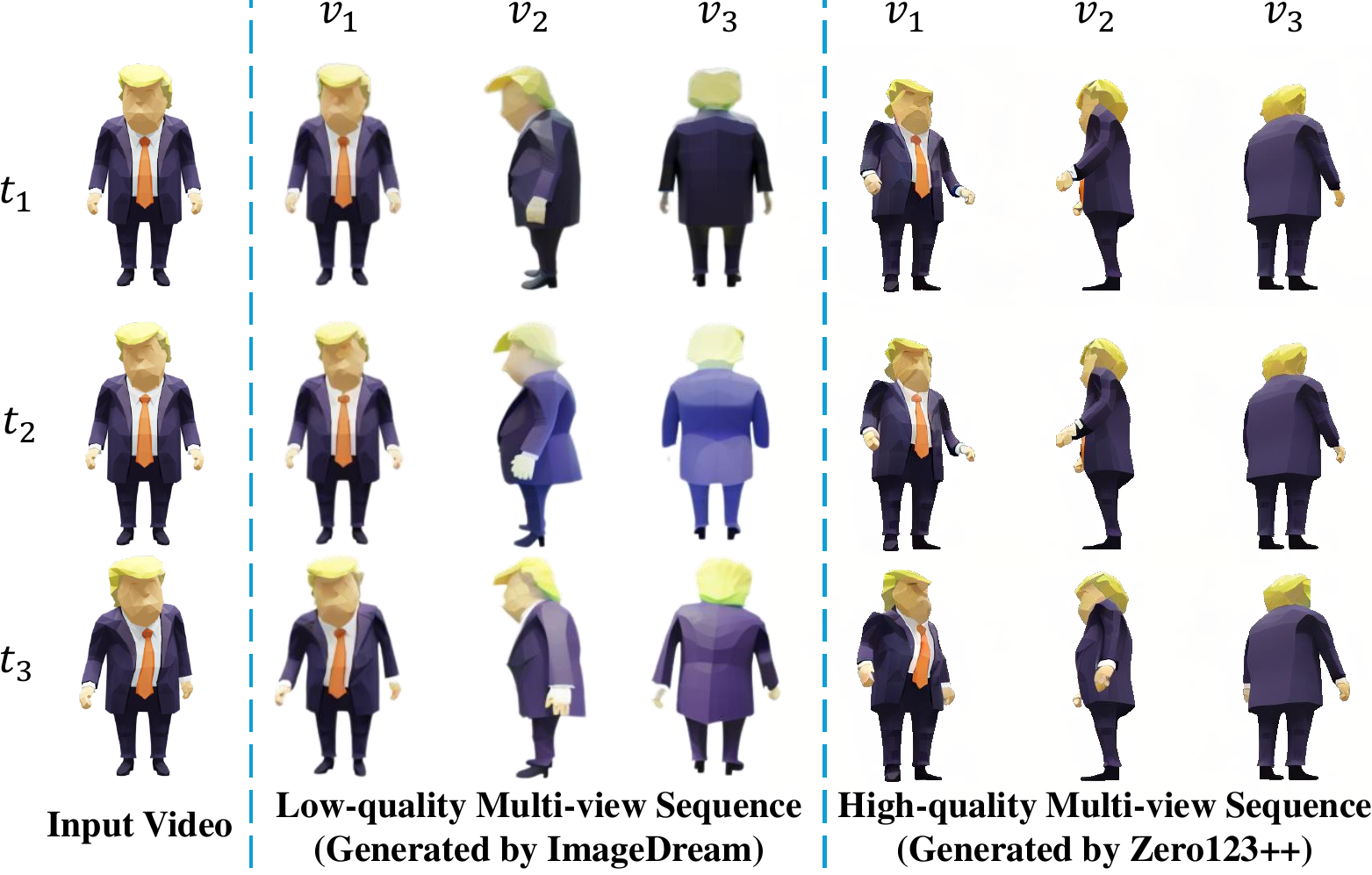}
\caption{Example on low-quality images generated by ImageDream and high-quality images generated by Zero123++, both using the same input video.}\label{imagedream}
\end{figure}

\begin{table*}[t]
	\centering
	\renewcommand\arraystretch{1.}
	\scalebox{0.85}{
	\begin{tabular}{c|cccc|cccc}
        \Xhline{1.5px}
		\multirow{2}{*}{\textbf{Methods}} & \multicolumn{4}{c|}{\textbf{High-quality Input Images}} & \multicolumn{4}{c}{\textbf{Low-quality Input Images}} \\ \cline{2-9}
          &\multicolumn{1}{c}{\textbf{CLIP $\uparrow$}} & \multicolumn{1}{c}{\textbf{LPIPS $\downarrow$}} & \multicolumn{1}{c}{\textbf{FVD $\downarrow$}} & \multicolumn{1}{c|}{\textbf{FID-VID $\downarrow$}} & \multicolumn{1}{c}{\textbf{CLIP $\uparrow$}} & \multicolumn{1}{c}{\textbf{LPIPS $\downarrow$}} & \multicolumn{1}{c}{\textbf{FVD $\downarrow$}} & \multicolumn{1}{c}{\textbf{FID-VID $\downarrow$}}
         \\ \hline
        \textbf{STAG4D \cite{zeng2024stag4d}}  & 0.9078 & 0.1354 & 986.8271 & 26.3705 &  0.9026 & 0.1437 & 1311.8770 & 40.8664  \\        
        \hline
        \textbf{DS4D-GA (Ours)} & 0.9206 & 0.1311 & 799.9367 & 26.1794 & 0.9195 & 0.1380 & 849.8154 & 25.8726 \\
        \textbf{DS4D-DA (Ours)} & \textbf{0.9225} & \textbf{0.1309} & \textbf{784.0235} & \textbf{24.0492} & \textbf{0.9221} & \textbf{0.1339} & \textbf{805.4721} & \textbf{24.0623} 
	    \\ \Xhline{1.5px}
	\end{tabular}}
	\caption{Evaluation and comparison of the performance when facing low-quality input images and high-quality input images. The best score is highlighted in bold.}
\label{robustness_result}
\end{table*}
\section{Training Details on Discussion}
In this section, we present the training details of experiments about Discussion (Section 5) in our manuscript. For a fair comparison, the training settings are the same as 4D-GS \cite{wu20244d}.

\begin{figure*}[]%
\centering
\includegraphics[width=1.0\textwidth]{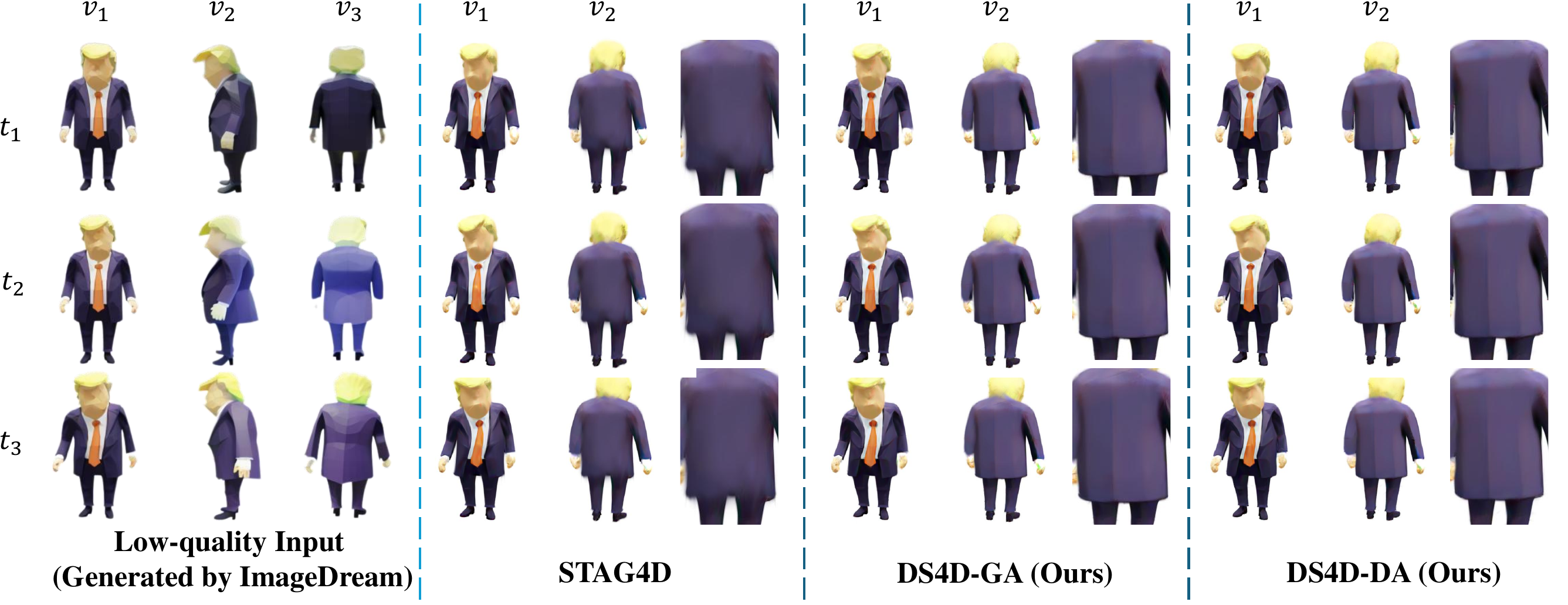}
\caption{Qualitative comparison on 4D generation results with low-quality inputs. For each method, we render results under two novel views at three timestamps.}\label{imagedream_results}
\end{figure*}

\begin{figure}[h]%
\centering
\includegraphics[width=0.4\textwidth]{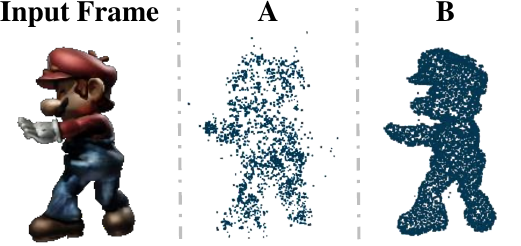}
\caption{Comparison regarding whether using point clouds generated by a large reconstruction model as initialization. A: The model without point initialization. B: The model with point initialization.}\label{points}
\end{figure}

\textbf{Network Architecture.} We add our proposed DSFD in 4D-GS. Meanwhile, we directly replace the spatial-temporal structure encoder and multi-head Gaussian deformation decoder of 4D-GS with our TSSF. Since the multi-view sequences from Neu3D's dataset are authentic, we use TSSF-GA instead of TSSF-DA. Additionally, the hyperparameter settings of HexPlane are the same as 4D-GS. In detail, the basic resolution of HexPlane is 64, and is unsampled by 2 and 4.

\textbf{Training Settings.} We use the same dense point cloud generated by SFM as 4D-GS. Then we also downsample point clouds lower than 100k. The learning rate of TSSF and deformation MLP are set to $1.6\times 10^{-3}$ which decreases to $1.6\times 10^{-5}$. The batch size is 1. Following 4D-GS, we do not use opacity reset operation. The overall training iterations spend 14000 for each scene.

Note that: a) Our DS4D includes point initialization (init.) via LRM. However, in Neu3D's data, common point cloud init. (e.g., 4D-GS) uses colmap rather than LRM. Thus, it is unfair to compare DS4D and 4D-GS due to the different init.. b) DSFD and TSSF are core contributions, thus inserting them into 4D-GS can directly demonstrate the effectiveness of our contributions in real-world scenes.

\section{More details on Datasets}
In this section, we present more details on \textbf{four datasets}.

\textbf{Consistent4D Dataset.} Following \cite{zeng2024stag4d}, we use seven 30-frame video in front view as input video, and their ground-truth with four novel views (azimuth angles of $-75^{\circ}$, $15^{\circ}$, $105^{\circ}$, and $195^{\circ}$, respectively) as evaluation. 

\textbf{Objaverse Dataset.} We random sample seven dynamic objects from \cite{deitke2023objaverse,liang2024diffusion4d}. The 24-frame ground truth under 360$^\circ$ cameras (the range of azimuth angles is $[0^{\circ}, 360^{\circ}]$) rendered from each object is used as evaluation. Meanwhile, we use seven 24-frame videos in front view as input video. Compared to Consistent4D Dataset, the object in Objaverse Dataset has more complex motion, e.g., suddenly waving at some time.

\textbf{Neu3D's Dataset.} We use three real-world scenarios from Neu3D's dataset \cite{li2022neural}. Each scene has 300 frames with 20 cameras, a total of 6000 high-quality images. Following \cite{wu20244d}, we use 300 frames under the front camera pose as evaluation, others are used as training videos.

\textbf{Data from Online Sources.} Following \cite{zeng2024stag4d}, we introduce some challenging videos from online sources for qualitative evaluation. Moreover, we also generate input videos by Stable Video Diffusion \cite{blattmann2023stable}. Each input video has 14 or 30 frames.

\section{Experiments}
In this section, we conduct more experiments to evaluate our method.

\subsection{Robustness of Our Methods} \label{robust}
In this section, we explore whether the quality of input images has a huge influence on the generation results of our methods. 

Specifically, we construct a dataset with low-quality input multi-view images. Using the same input video as Tab.1 of the main manuscript, we leverage ImageDream \cite{wang2023imagedream} to produce a series of multi-view sequences. Then, we select multi-view images with inconsistency or shape deformation from the generation multi-view sequences. These low-quality multi-view images are grouped as the low-quality input images. The example data of low-quality inputs can be seen in Fig.~\ref{imagedream}. The low-quality inputs has serious inconsistency between different timestamps and has color fading and texture blurry compared with the high-quality inputs we used in Tab.1 of the main manuscript (e.g., the shape and color of the suit). 

Based on the low-quality inputs, we compare our methods with STAG4D. The quantitative and qualitative results are shown in Tab.~\ref{robustness_result} and Fig.~\ref{imagedream_results}. Our methods when using low-quality inputs maintain a similar performance compared to our results when using high-quality inputs. However, in image metrics (LPIPS) and video metrics (FID and FID-VID), STAG4D when using low-quality inputs has a significantly worse performance compared to STAG4D using high-quality inputs. Furthermore, in Fig.~\ref{imagedream_results}, STAG4D generates the blurry textures in the back view. The reason is that back's texture details are blurry in the low-quality input at some timestamps (e.g., at $t_1$ of $v_3$). In contrast, our methods can generate results with clear textures. This indicates that our methods can handle the low-quality inputs better than STAG4D. Since our methods decouple dynamic-static information at the feature-level. Even though input low-quality data, thanks to the robustness feature extraction ability of DINOv2, we can still leverage the inherent differences between features to decouple. The differences include the change of motion, shape and textures between the corresponding two frames.

In summary, the above experiments demonstrate that the quality of input images has few influence on the generation results of our methods and verify the robustness of our methods when input low-quality data.

\subsection{More Analysis on Ablation Experiments}
\begin{figure}[t]%
\centering
\includegraphics[width=0.4\textwidth]{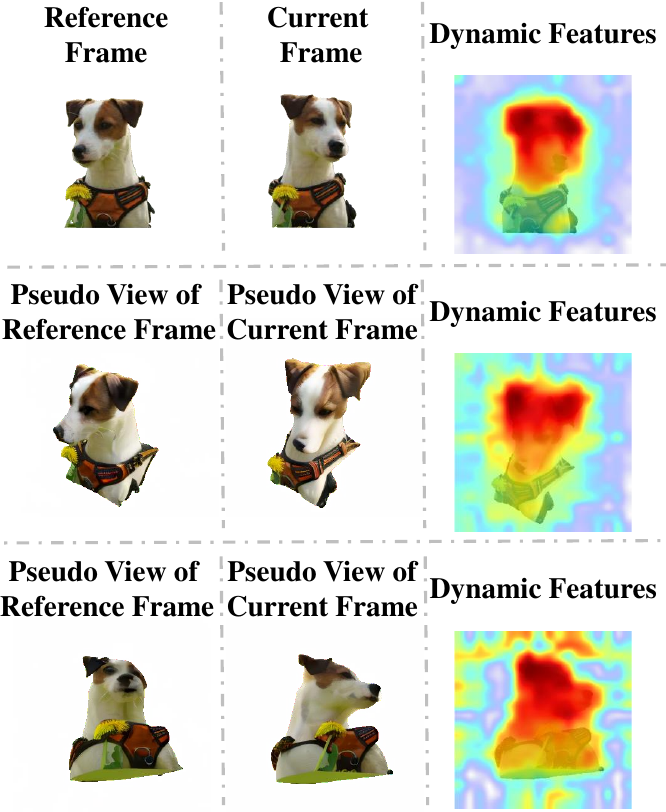}
\caption{Visualization on the heatmap of dynamic features in DSFD. The red region highlights the primary zone of interest in dynamic features.}\label{decouple}
\end{figure}

\begin{figure}[t]%
\centering
\includegraphics[width=0.45\textwidth]{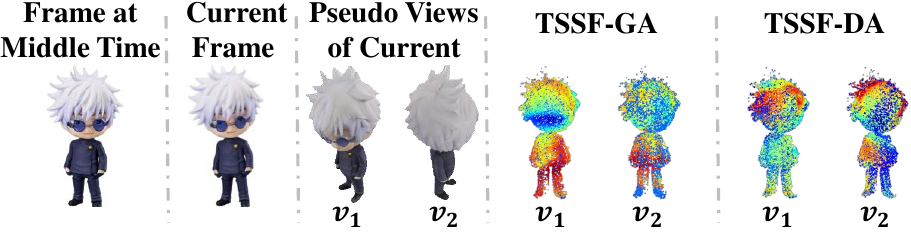}
\caption{Visualization on the score map of point features in TSSF-GA and TSSF-DA. The red area indicates model's high attention on dynamic information in point features of a specific view.}\label{fusion}
\end{figure}

\textbf{The effect on points initialization.} To validate the effect of point initialization, we add the point initialization to the baseline model, which is labeled as B (as shown in Table.2 of the manuscript). After performing point initialization, the performance of B model has improved on all metrics. Additionally, we visualize the Gaussian points from baseline model (labeled as A, as shown in Table.2 of the manuscript) and B in Fig.\ref{points}. Obviously, the Gaussian points of B are relatively uniform and denser in distribution than those of A, which ensures the stability of optimization and fidelity of the motion and shape fully learned by the model.

\textbf{The effect on LPIPS Loss.} To validate the effect of LPIPS Loss, we add the LPIPS Loss based on model B, labeled as C (as shown in Table.2 of the manuscript). The performance of C model has improved on all metrics. It indicates the effectiveness of LPIPS loss.

\subsection{More Visualization}
In this section, we provide more visualizations of features in DSFD and TSSF, respectively.

\textbf{DSFD.} In Fig.\ref{decouple}, we supplement more heatmaps of dynamic features obtained by DSFD decoupling features from the current and reference frame features. The red area indicates the primary region of interest in the features. The dog's head movements indicate the main motion trends between the reference frame and the current frame. No matter what kind of novel views, dynamic features decoupled by our DSFD can accurately represent the dynamic zones. It once again demonstrates our method can acquire accurate dynamic features using DSFD.

\begin{figure}[h]%
\centering
\includegraphics[width=0.45\textwidth]{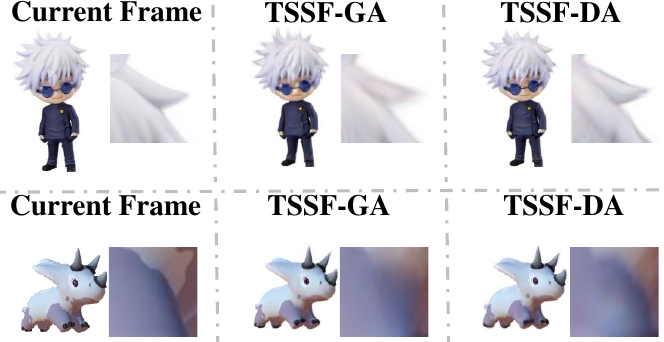}
\caption{Qualitative comparison between model using TSSF-GA and model using TSSF-DA at current timestamp. Their corresponding visualization on the score map of point features is as shown in Fig.\ref{fusion} and Figure.8 (b) in our manuscript.}\label{vis_fusion}
\end{figure}

\textbf{TSSF.} In Fig.\ref{fusion}, we supplement more score maps of selecting similarity dynamic information from point features of different views by TSSF-GA and TSSF-DA. Moreover, Fig.\ref{vis_fusion} shows the corresponding generation results at the current timestamp, including the same example as Figure.8 (b) in the manuscript and the same example in Fig.\ref{fusion}.

Specifically, the head movements of the person indicate the primary trend in motion between the middle and current time. The red area indicates the model's high attention on dynamic information in point features of a specific view. According to score maps based on different views, two approaches can capture a certain degree of similar dynamic information from different viewpoints. TSSF-GA is interested in both body and head in $v_1$, but TSSF-DA pays more concentration to the head. This is because TSSF-DA reduces the impact of novel views, resulting in TSSF-DA focusing more on capturing regions with a motion trend that is more similar to the front frame in other novel views. Thus, compared to TSSF-GA, TSSF-DA can produce results with clear details in the corresponding regions. For example, the hair texture of person at the top of Fig.\ref{vis_fusion}, and the leg texture of triceratops at the down of Fig.\ref{vis_fusion}. It once again indicates TSSF-DA can alleviate issues caused by novel views.

\begin{figure*}[]
    \centering
	\begin{minipage}{0.75\linewidth}
        \centerline{\includegraphics[width=1\linewidth]{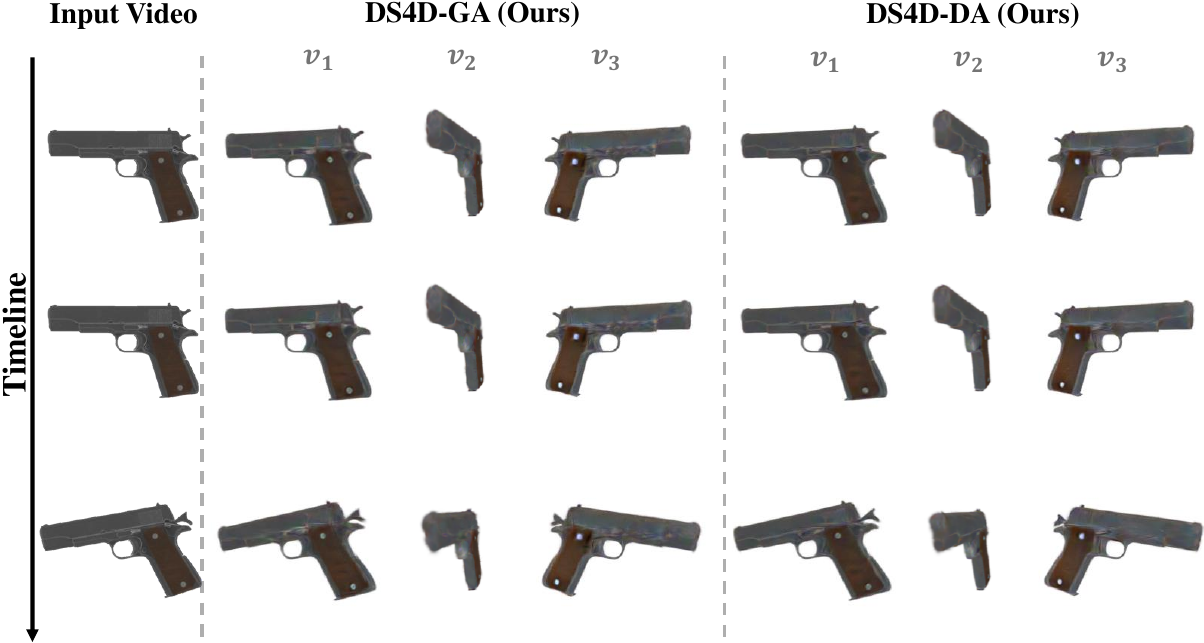}}
	\end{minipage}
	\\ \vspace{0.5mm}
	\begin{minipage}{0.75\linewidth}
        \centerline{\includegraphics[width=1\linewidth]{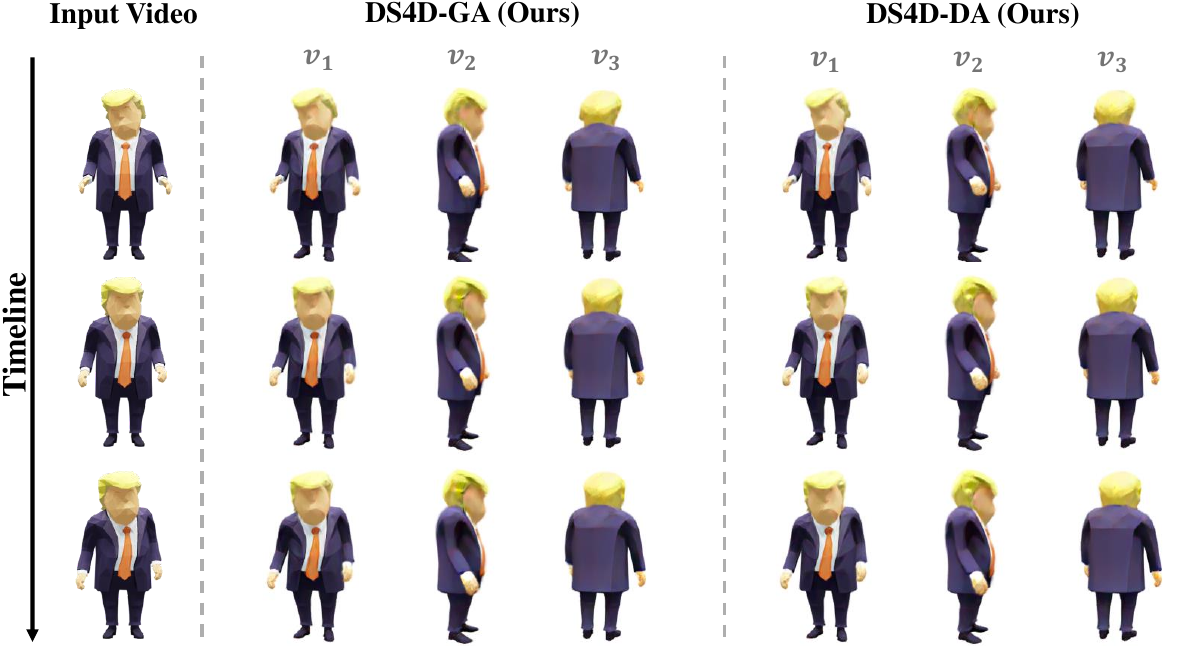}}
	\end{minipage}
	\\ \vspace{0.5mm}
	\begin{minipage}{0.75\linewidth}
        \centerline{\includegraphics[width=1\linewidth]{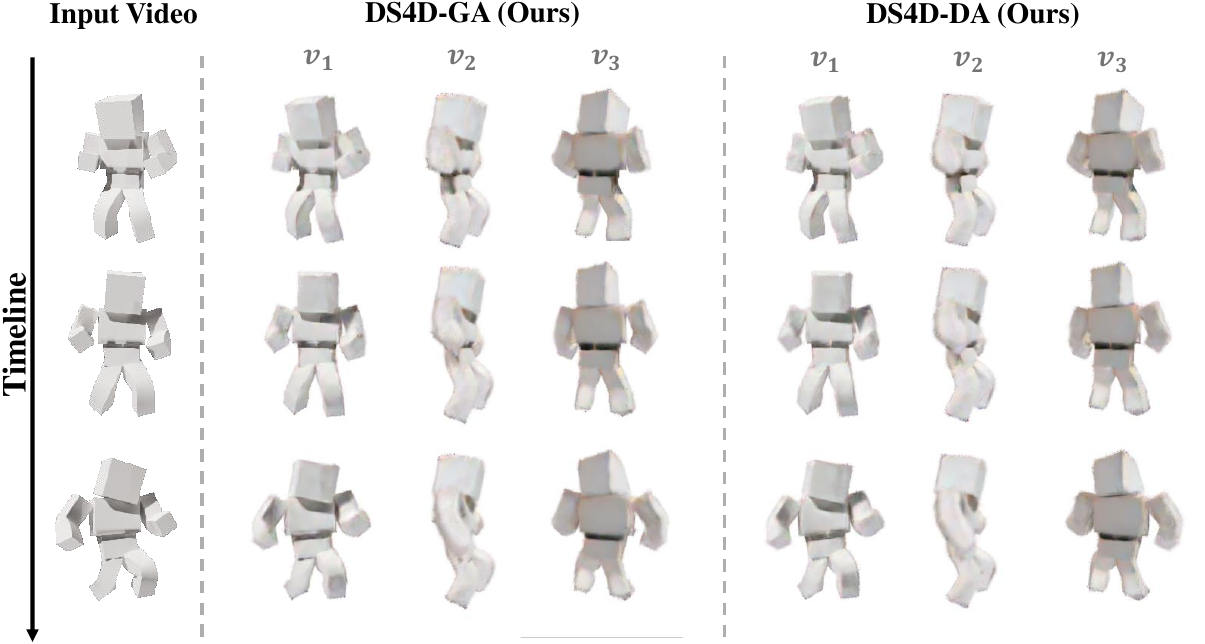}}
	\end{minipage}
	\caption{More Results for 4D Generation using DS4D-GA and DS4D-DA.}
\label{show_results}
\end{figure*}

\begin{figure*}[t]
    \centering
	\begin{minipage}{1\linewidth}
        \centerline{\includegraphics[width=1\linewidth]{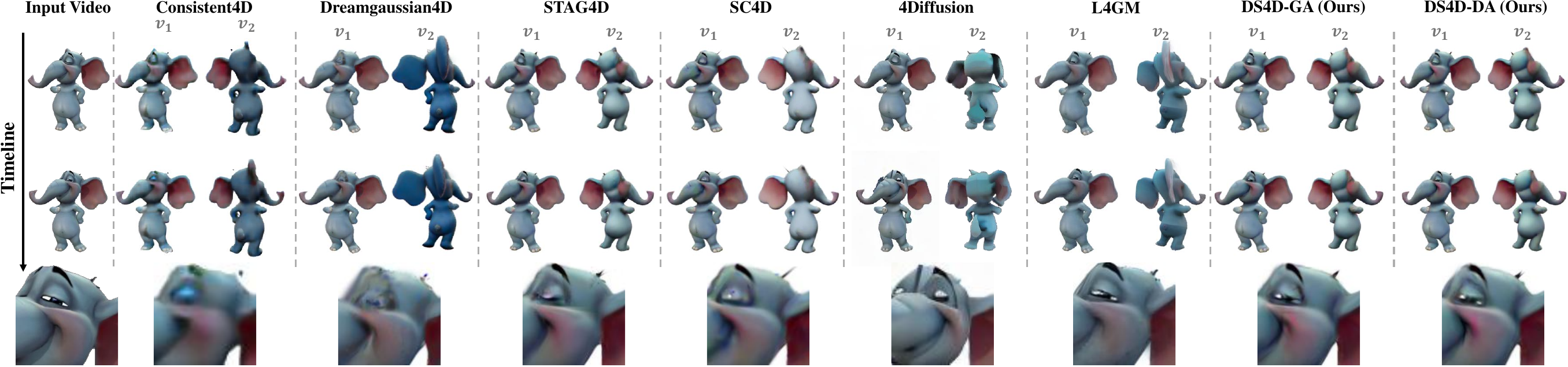}}
	\end{minipage}
	\\ \vspace{1mm}
	\begin{minipage}{1\linewidth}
        \centerline{\includegraphics[width=1\linewidth]{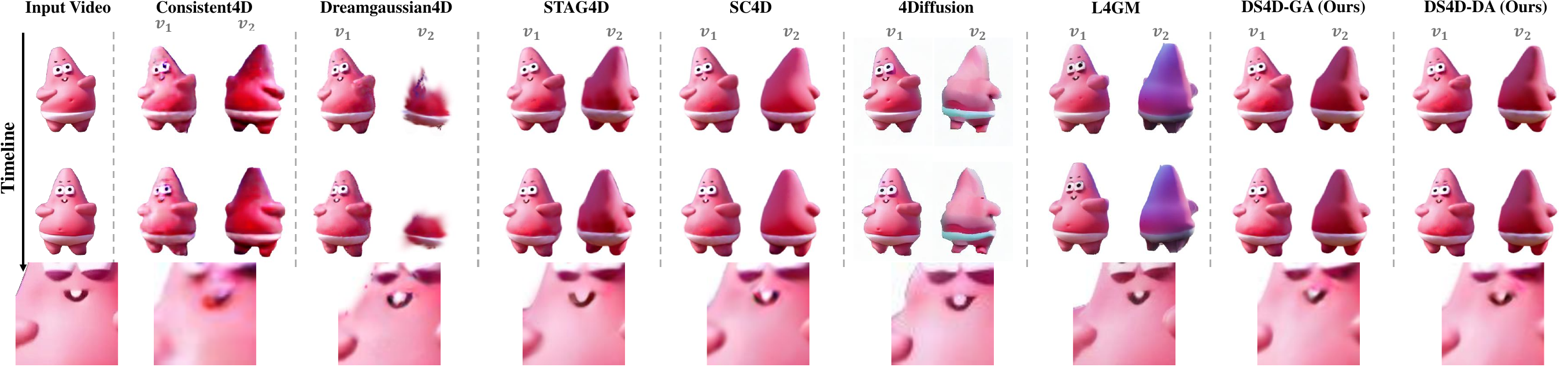}}
	\end{minipage}
	\caption{Qualitative comparison on video-to-4D generation. For each method, we render results under two novel views at two timestamps.}
\label{compare}
\end{figure*}

\begin{figure*}
  \centering
  \begin{subfigure}{0.48\linewidth}
    \scalebox{1.}{
    \includegraphics[width=1\linewidth]{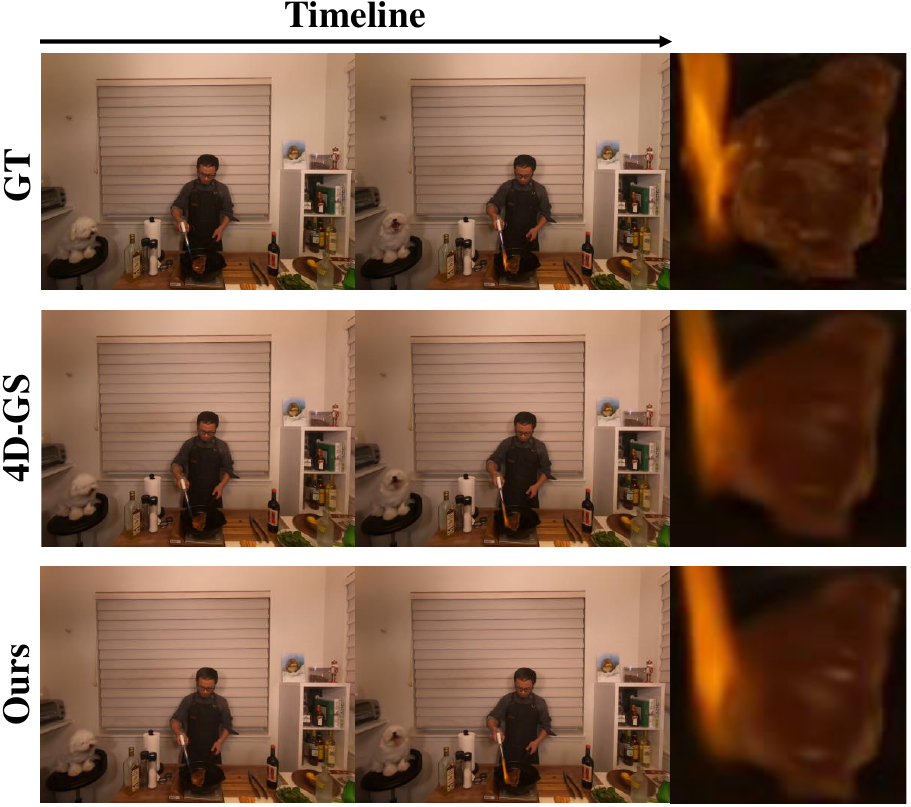}}
    \caption{Sear steak}
    \label{s1}
  \end{subfigure}
  \begin{subfigure}{0.48\linewidth}
   \scalebox{1.}{
    \includegraphics[width=1\linewidth]{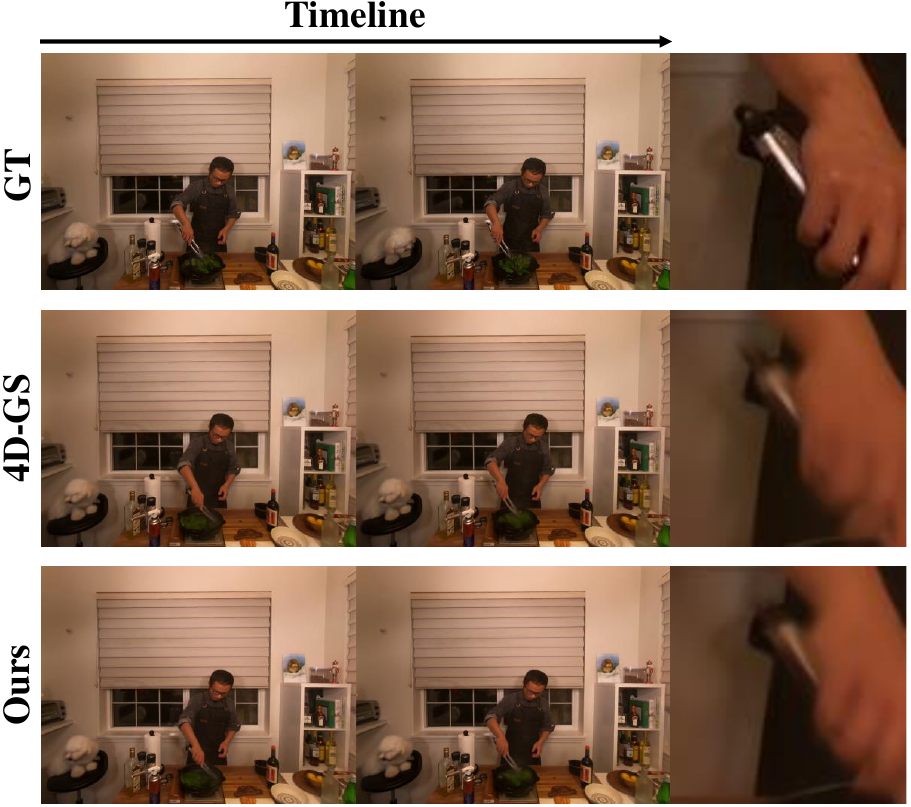}}
    \caption{Cook spinach}
    \label{s2}
  \end{subfigure}
  \caption{Visualization of real-world 4D scene generation compared with 4D-GS.}
  \label{discussion}
\end{figure*}

\subsection{More Qualitative Comparisons}

We supplement more qualitative comparisons in Fig.\ref{compare}. We compare our methods DS4D-GA and DS4D-DA with other SOTA methods, including Consistent4D \cite{jiang2024consistentd}, Dreamgaussian4D \cite{ren2023dreamgaussian4d}, STAG4D \cite{zeng2024stag4d}, SC4D \cite{wu2024sc4d}, 4Diffusion \cite{zhang20244diffusion}, and L4GM \cite{ren2024l4gm}. All the experiments of the methods are carried out using the code from their official GitHub repository.

\subsection{More Examples on 4D Content Generation}

We supplement more 4D content generation examples produced by DS4D-GA and DS4D-DA in Fig.\ref{show_results}.

\subsection{More Examples on Real-World Scenario}

We supplement more real-world 4D scene generation examples using our method and baseline 4D-GS \cite{yang2023real} in Fig.\ref{discussion}. The experiments of 4D-GS are carried out using the code from their official GitHub repository.

\subsection{Time Consuming on Decoupling}
As mentioned in our manuscript, direct decoupling always costs considerable computation time. To intuitively evaluate the time-consuming advantage of our decoupling approach compared to direct decoupling, we provide the running time of each approach on decoupling dynamic-static features from a frame features with a 30-frame video, as shown in Tab.\ref{time}. Undoubtedly, our decoupling approach is about 14 times faster than direct decoupling.  
\begin{table}[]
	\centering
	\renewcommand\arraystretch{1.2}
	\scalebox{1}{
    \begin{tabular}{c|c}
	    \hline
	    \multicolumn{1}{c}{\textbf{Direct Decoupling}}&
		\multicolumn{1}{|c}{\textbf{Our Decoupling Approach}} \\ \hline
		118.326 (ms) & 8.241 (ms) \\ \hline
	\end{tabular}}
	\caption{Comparison of running time with different decoupling approaches. All approaches are tested on a NVIDIA 3090 GPU.}
\label{time}
\end{table}

\section{Limitations}
Limited to the resolution of input video, it is challenge for our method to produce high-resolution 4D contents , e.g., over 2K resolution.

\newpage

{
    \small
    \bibliographystyle{ieeenat_fullname}
    \bibliography{main}
}

\end{document}